\documentclass{article}
\usepackage[utf8x]{inputenc}
\usepackage[square, numbers, ]{natbib}
\usepackage[shortlabels]{enumitem}
\usepackage{enumitem}

\usepackage{ifpdf} 
\usepackage{graphicx}

\usepackage{todonotes}
\usepackage[utf8x]{inputenc}
\usepackage[square, numbers]{natbib}
\usepackage[shortlabels]{enumitem}
\usepackage{enumitem}
\usepackage{tabularx}
\usepackage{amsmath, amsfonts}
\usepackage{tikz}
\usetikzlibrary{decorations.pathreplacing,calligraphy}
\usepackage{subcaption}
\usepackage{xcolor}
\usepackage{bigdelim}
\usepackage[labelsep=period]{caption}
\usepackage{ragged2e}
\usepackage{blindtext}
\usepackage{float}
\usepackage{tikz-network}
\usetikzlibrary{calc,arrows,positioning,shapes.arrows}
\usepackage{multicol}
\usepackage{multirow}
\newcommand{\beq}{\begin{equation}}
\newcommand{\eeq}{\end{equation}}

\usepackage{ifpdf}
\usepackage{graphicx}

\usepackage{chngcntr}
\counterwithout{figure}{section}

\begin{document}
\bibliographystyle{IEEEtranN}

\title{Anticipatory Understanding of Resilient Agriculture to Climate}
\author{Dr. David E. Willmes \and Nick S. Krall \and Dr. James H. Tanis \and Dr. Zachary Terner \and Dr. Fernando T. Tavares \and Dr. Chris W. Miller \and Joe Haberlin III \and Matt Crichton \and Dr. Alexander Schlichting
}

\maketitle

\begin{abstract}
With billions of people facing moderate or severe food insecurity, the resilience of the global food supply will be of increasing concern due to the effects of climate change and geopolitical events. In this paper we describe a framework to better identify food security hotspots using a combination of remote sensing, deep learning, crop yield modeling, and causal modeling of the food distribution system. While we feel that the methods are adaptable to other regions of the world, we focus our analysis on the wheat breadbasket of northern India, which supplies a large percentage of the world’s population. We present a quantitative analysis of deep learning domain adaptation methods for wheat farm identification based on curated remote sensing data from France. We model climate change impacts on crop yields using the existing crop yield modeling tool WOFOST and we identify key drivers of crop simulation error using a longitudinal penalized functional regression. A description of a system dynamics model of the food distribution system in India is also presented, along with results of food insecurity identification based on seeding this model with the predicted crop yields.
\end{abstract}

\section{Introduction}
In 2022, approximately 2.4 billion people faced moderate or severe food insecurity, representing almost 30 percent of the global population~\cite{FAO2023}.
This problem has been exacerbated recently due to the war in Ukraine as well as an increase in heat and droughts in some of the breadbaskets around the world. In India, the 2022 heat wave severely curtailed the country's wheat production, prompting India's Minister of Agriculture to publish a memo suggesting the need for an end-to-end system to monitor crop production and the delivery of grain to their citizens~\cite{Kumar_2022}.

We are focusing this research on the wheat breadbasket of northern India, which is responsible for sustaining not only the most populated nation on Earth, but also has become a significant source of grain to the rest of the world after the war in Ukraine  curtailed Ukrainian exports. While there are about 140 million farms in India, with about 40\% growing wheat in a given year, prediction of the amount of food available is difficult. Most of India's wheat crops are grown on small, family-owned farms, with an average size of about 3 acres according to the India Agricultural Census, and many different strains of wheat are grown in each of the wheat-producing states~\cite{indiacensus}. This creates an uncertainty in crop yield that necessitates an automated approach to crop identification and yield. 

In this paper, we will discuss the technical development of a framework to identify wheat fields using satellite remote sensing and deep learning classification, predict crop production under various climate scenarios using physics-based crop yield simulations, model the distribution of grain to food insecure populations using a system dynamics approach, and suggest potential courses of action that can improve the resiliency of the food system. An interactive dashboard in which a user can explore the implications of different courses of action has been built, which shows the results of running the system dynamics models for each region of interest.

While the focus of this research is on the wheat breadbasket of northern India, there are similarities to other scholarly work where techniques were developed for combining AI with remote sensing for agriculture, for example the recent work by Nakalembe and Kerner and their analysis of agriculture in sub-Saharan Africa~\cite{nakalembe2023considerations}. For instance, both regions suffer from a lack of ground truth for agricultural production, which makes it difficult to validate performance of the algorithms. In our case, concentrate our deep learning analysis on wheat production in France, where curated remote sensing datasets exist, and use domain adaptation techniques to transfer results to ou region of interest. The distribution and market dynamics portion of this research focuses on modeling the unique aspects of the Indian economy and governance.


\section{Remote Sensing-based Crop Identification}
In order to accurately forecast yield trends, the total growth area for each crop type must be assessed in an automated manner.
While there are several options for obtaining georeferenced crop type labels at scale, remote sensing data was chosen due to its potential for timely and scalable collection of rich spatial, spectral, and temporal information necessary for accurate crop type identification.
With the recent growth of machine learning (ML) capabilities applied to computer vision and remote sensing, we selected a common family of ML architectures that have shown to perform well on the crop type mapping applications.
These architectures include semantic segmentation convolutional neural networks (CNNs) which can exploit spatial relationships in pixels, long short-term memory recurrent neural networks (LSTMs) that can accurately capture temporal patterns, and combinations of the two to jointly handle spatial, spectral, and temporal information.

\subsection{Data}
Several satellite constellations were considered and the European Space Agency’s (ESA) Sentinel-2 (S2) satellite was selected.
The S2 constellation was selected because of its mix of medium-high spatial resolution at multiple spectral bands (shown in Table~\ref{tab:s2_band_info}), medium-high revisit rate of five days, and availability of free data.
An automated data download system was created to pull S2 imagery from an AWS hosting service and store on a local network attached storage drive, then indexed in a geospatially enabled SQL database. Python was used to implement all methods and processing.

\begin{table}[]
  \caption{Information on the 13 bands in Sentinel-2 satellite sensors.}
  \label{tab:s2_band_info}
  \begin{tabular}{| l | c | c |}
  \hline
  Sentinel-2 Spectral Band & Central Wavelength (µm) & Resolution (m) \\
  \hline
  Band 1: Coastal aerosol  & 0.443                   & 60             \\
  Band 2: Blue             & 0.490                   & 10             \\
  Band 3: Green            & 0.560                   & 10             \\
  Band 4: Red              & 0.665                   & 10             \\
  Band 5: Red Edge 1       & 0.705                   & 20             \\
  Band 6: Red Edge 2       & 0.740                   & 20             \\
  Band 7: Red Edge 3       & 0.783                   & 20             \\
  Band 8: NIR              & 0.842                   & 10             \\
  Band 8A: Red Edge 4      & 0.865                   & 20             \\
  Band 9: Water vapor      & 0.945                   & 60             \\
  Band 10: SWIR 1          & 1.375                   & 60             \\
  Band 11: SWIR 2          & 1.610                   & 20             \\
  Band 12: SWIR 3          & 2.190                   & 20             \\
  \hline
  \end{tabular}
  \end{table}

One consideration for data preprocessing was whether to use bottom of atmosphere (BoA) or top of atmosphere (ToA) data.
The BoA product benefits from being atmospherically corrected, but as shown in a recent crop type classification method \cite{breizhcrops}, ToA data can lead to equivalent accuracy.
ToA data (S2 processing level of L1C) was used in our models due to ease of adoption of methods to other regions where atmospherically corrected data is not possible.

\begin{figure}[H]
  \centering
  \includegraphics[width=8cm]{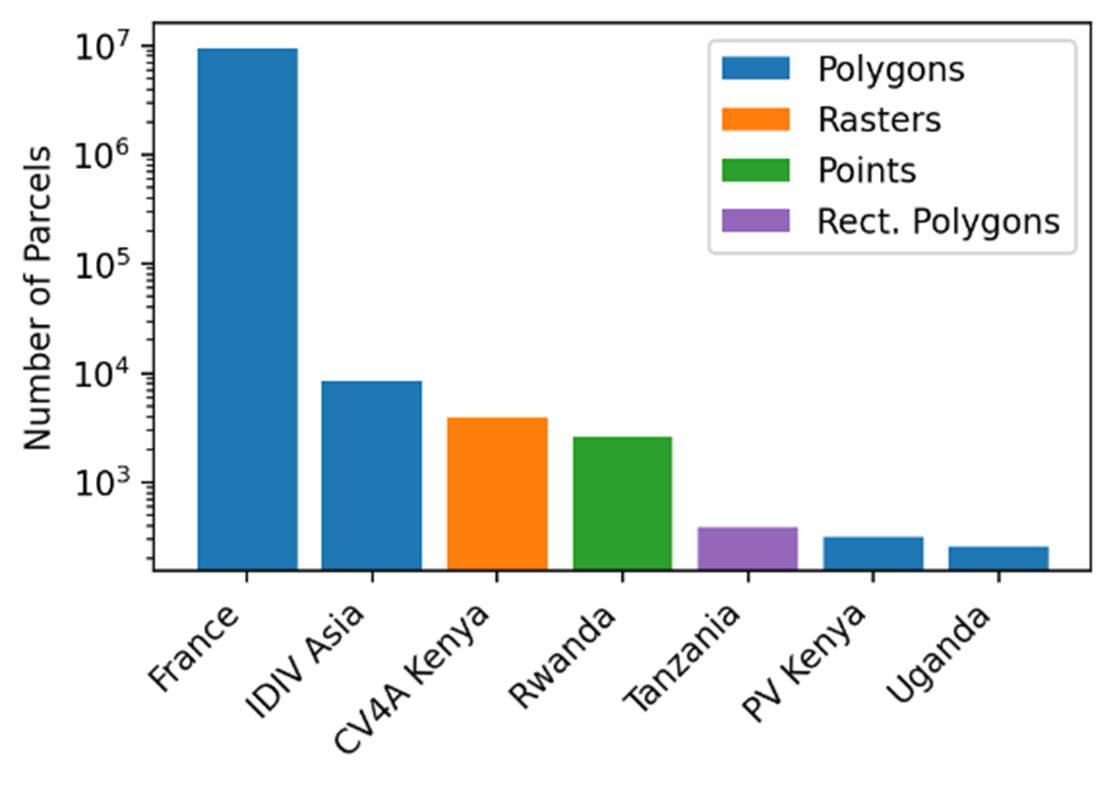}
  \caption{Crop type classification datasets are spread across many domains and formats, requiring specialized preprocessing to homogenize data for consistent usage. }\label{fig:datasets_formats}
\end{figure}

Robust crop type classification across different climates and terrains ideally involves labeled data from each unique bioclimate.
In order to efficiently use these unique datasets it is important to create globally consistent labels, but datasets from each region have different formats, as shown in Figure~\ref{fig:datasets_formats}.

\begin{figure}[H]
  \centering
  \includegraphics[width=8cm]{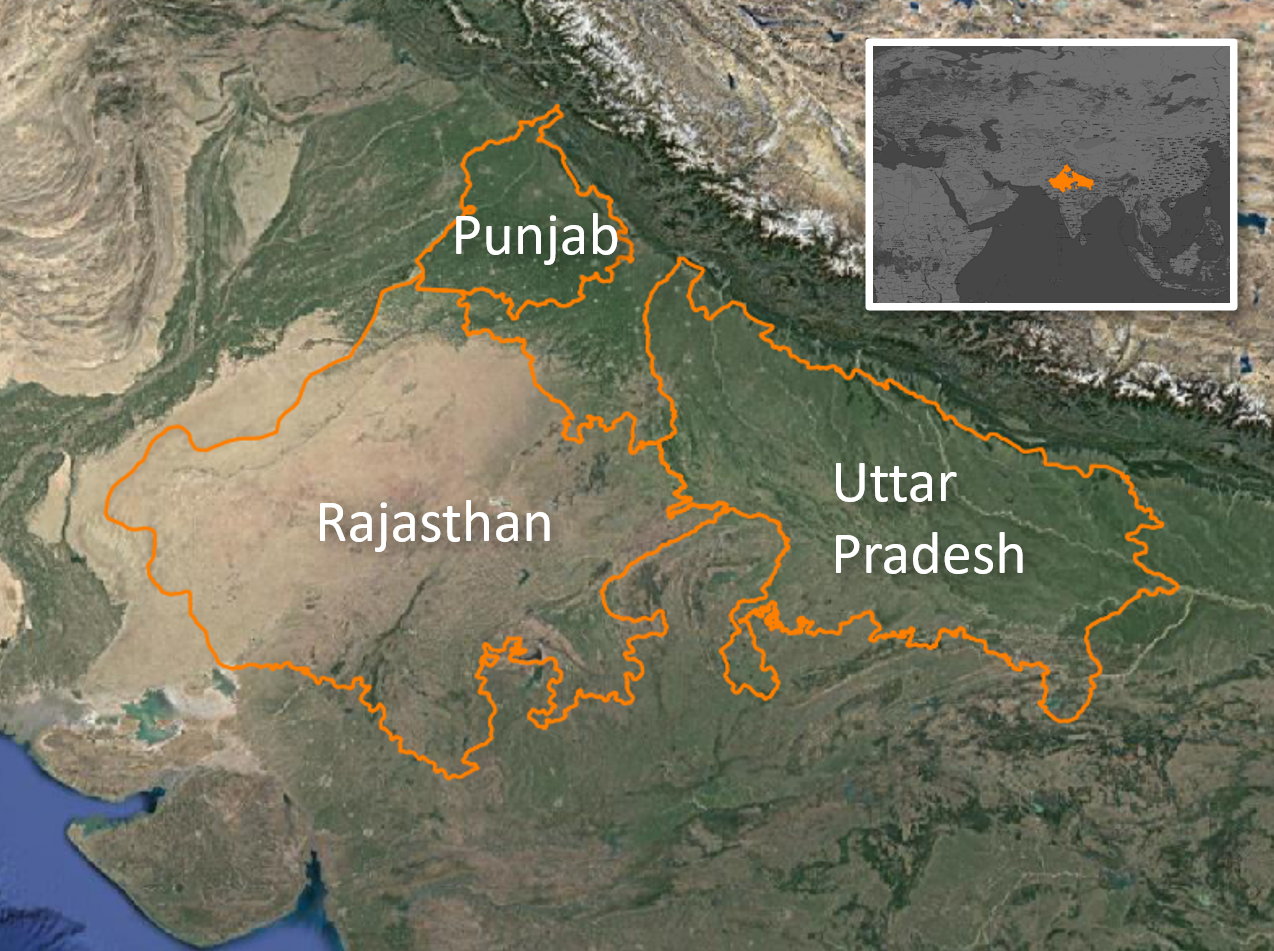}
  \caption{The three focus states in Northern India encompass several different geographies.}\label{fig:india_focus_states_map}
\end{figure}

Although the primary focus area for evaluating the overall food security approach is the northern India region, (Figure~\ref{fig:india_focus_states_map}), no known sufficient ground truth exists for crop type classification in these regions, as concluded from a significant literature review.
Since modern deep learning methods require large scale datasets, we train and validate our models on label-rich areas and then test the models in northern India.
This training strategy introduces challenges due to domain transfer which will be addressed in a later section.

\begin{figure}[H]
  \centering
  \includegraphics[width=8cm]{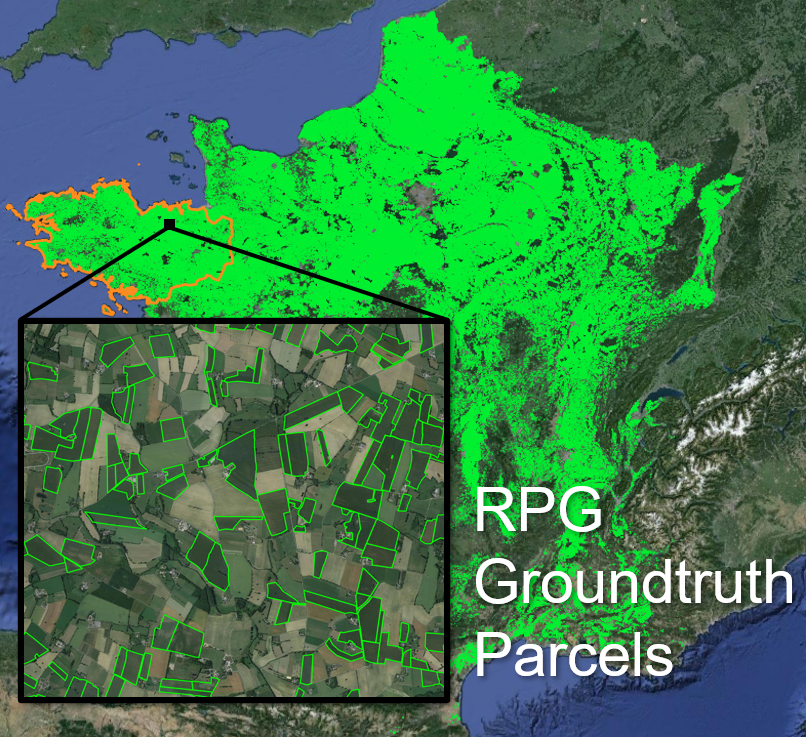}
  \caption{France's RPG dataset provides parcel-level crop type annotations for the full country.}\label{fig:france_rpg_parcels_with_zoom}
\end{figure}

Following \cite{breizhcrops}, we first train and evaluate our models in France. As shown in Figure~\ref{fig:france_rpg_parcels_with_zoom}, we use the associated ground truth parcels provided by the French National Institute of Forest and Geography Information as part of the EU’s Common Agricultural Policy.
This dataset is known as the Agricultural Land Parcel Information System (Registre Parcellaire Graphique or RPG) and consists of 328 unique crop labels grouped into 23 groups \cite{rpg_dataset}.
The crop class “winter wheat” is the primary focus of our study.

\begin{figure}[H]
  \centering
  \includegraphics[width=8cm]{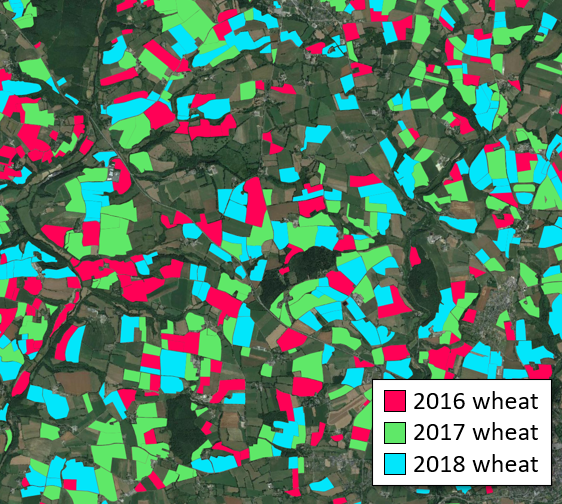}
  \caption{Wheat parcels rotate across years, deeming it necessary to use specific inidividual year's imagery and labels.}\label{fig:france_rpg_multi_years}
\end{figure}

We investigate growing periods in years 2017, 2018, and 2019. Unique RPG layers exist for each year. Since many farmers grow different crops throughout a period of several years, as shown in Figure~\ref{fig:france_rpg_multi_years}, it is necessary to use temporally appropriate labels. We also study the effect of using several different time periods and scales, and all methods incorporate multiple time steps to account for the full growth cycle of crops.

\subsubsection{Data Preprocessing}
\begin{figure}[H]
  \centering
  \includegraphics[width=10cm]{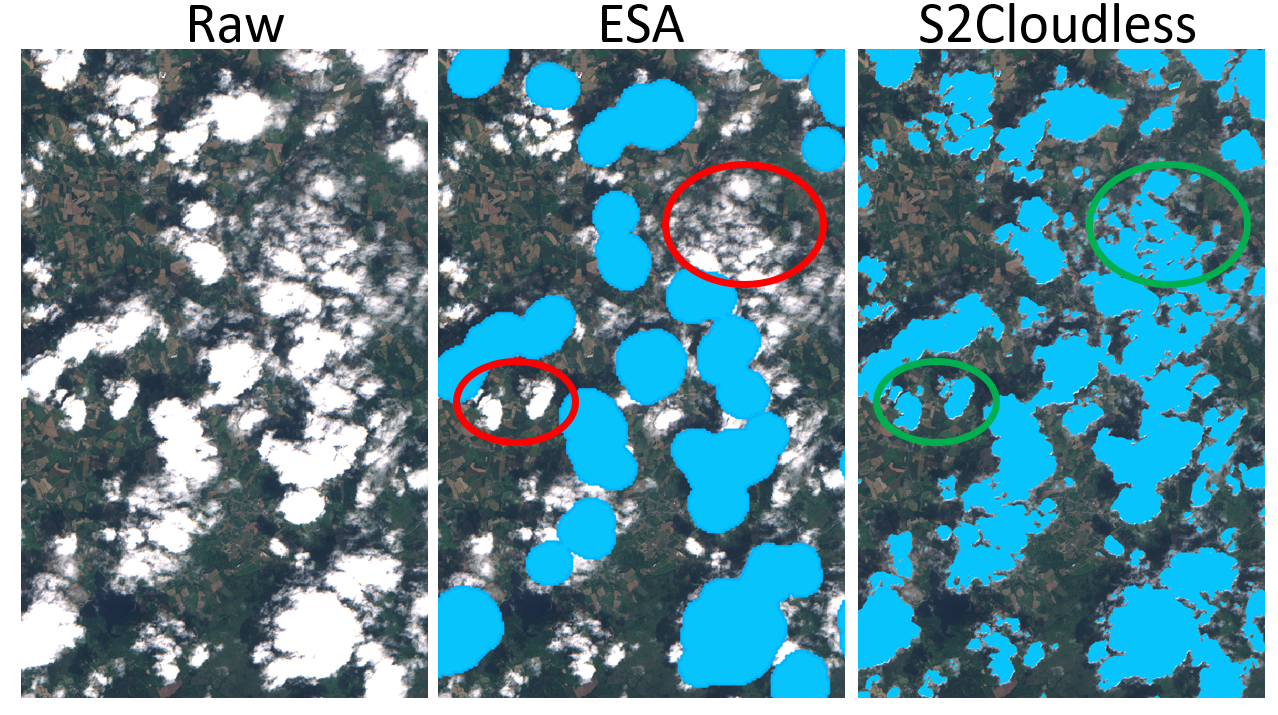}
  \caption{The tuned S2Cloudless model can capture most or all cloud instances and cloud types in its per-pixel mask, where the default Sentinel-2 vector cloud mask fails. The green ovals in S2Cloudless highlight the clouds that are successfully masked that are incorrectly missed in the red ovals of ESA. }
  \label{fig:cloud_mask_comparison}
\end{figure}

Once all data is downloaded for the desired locations and time periods, all bands are upsampled to the highest spatial resolution of 10 m to ensure uniform inputs to the model. 
Although ESA provides cloud masks associated with each collect, we determined the masks were not suitable for training applications due to missing some significant cloud groupings, circled in red in Figure~\ref{fig:cloud_mask_comparison}. As an alternative, we used a separate package called S2Cloudless \cite{skakun2022cloud}
and qualitatively tuned it on several different cloud types and regions within France. The resulting cloud mask is more complete than the ESA mask, with the correct masking highlighted in green in the figure.

\subsection{Crop Classification Models}
We investigate several methods that collectively use combinations of spatial, spectral, and temporal data.
In selecting models to use, there are two major considerations.
The first is model complexity, which includes the number of parameters, training stability, and computation time. Model complexity informs the amount and types of data that are required to obtain accurate results. The second consideration is model generalizability, or the ability of the model to perform similarly across multiple geographic, temporal, spectral, and cultural (economic, farming practices, etc.) domains.

The first model considered is spatial-spectral, which consists of a CNN encoder-decoder architecture (UNet \cite{unet} architecture and a ResNet50 \cite{resnet} encoder backbone) with a separate temporal merging step outside of the training process.
Multispectral images from the entire growth cycle are randomized and provided to the model during training, with the idea of generalizing the spatial and spectral factors to detect specific crop types at any time during the growth cycle.

\begin{figure}[H]
  \centering
  \includegraphics[width=12cm]{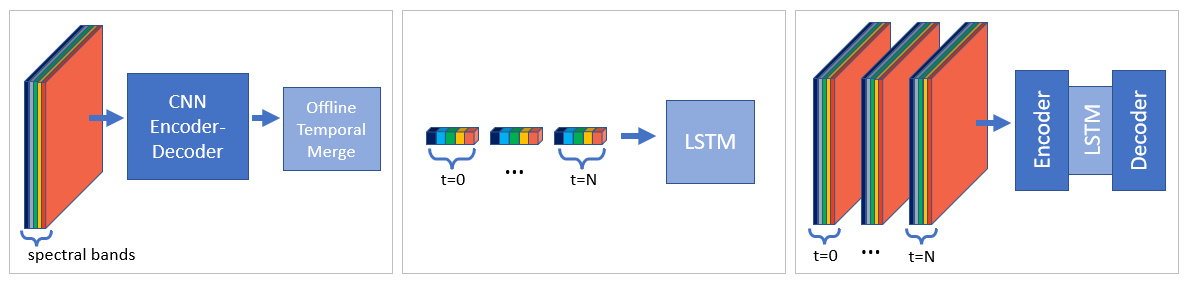}
  \caption{\textit{Left}: Spatial-Spectral (SS) model. \textit{Center}: Spectral-Temporal (ST) model. \textit{Right}: Spatial-Spectral-Temporal (SST) model.}\label{fig:three_model_types}
\end{figure}

The spatial-spectral model has been heavily researched in the broader computer vision community and has mature, successful architectures for various applications \cite{rs_model_survey}.
However, because this model type does not directly incorporate multiple time steps in training, it does not explicitly learn temporal growth characteristics, which aid in the crop type classification task. Figure~\ref{fig:three_model_types} shows the three model types investigated.

The second model type is spectral-temporal, which uses an LSTM recurrent neural network to explicitly learn temporal growth characteristics during training.
Each pixel is independently processed as a separate sample, so these are typically the slowest in inference of a large image.
However, because each pixel is a separate sample, it is also the most easily extensible to all ground truth vector formats including point, polygon, rectangular, and more.
One major drawback is that no contextual information is considered during training, so neighboring pixels do not explicitly inform the current pixel.

\begin{figure}[H]
  \centering
  \includegraphics[width=12cm]{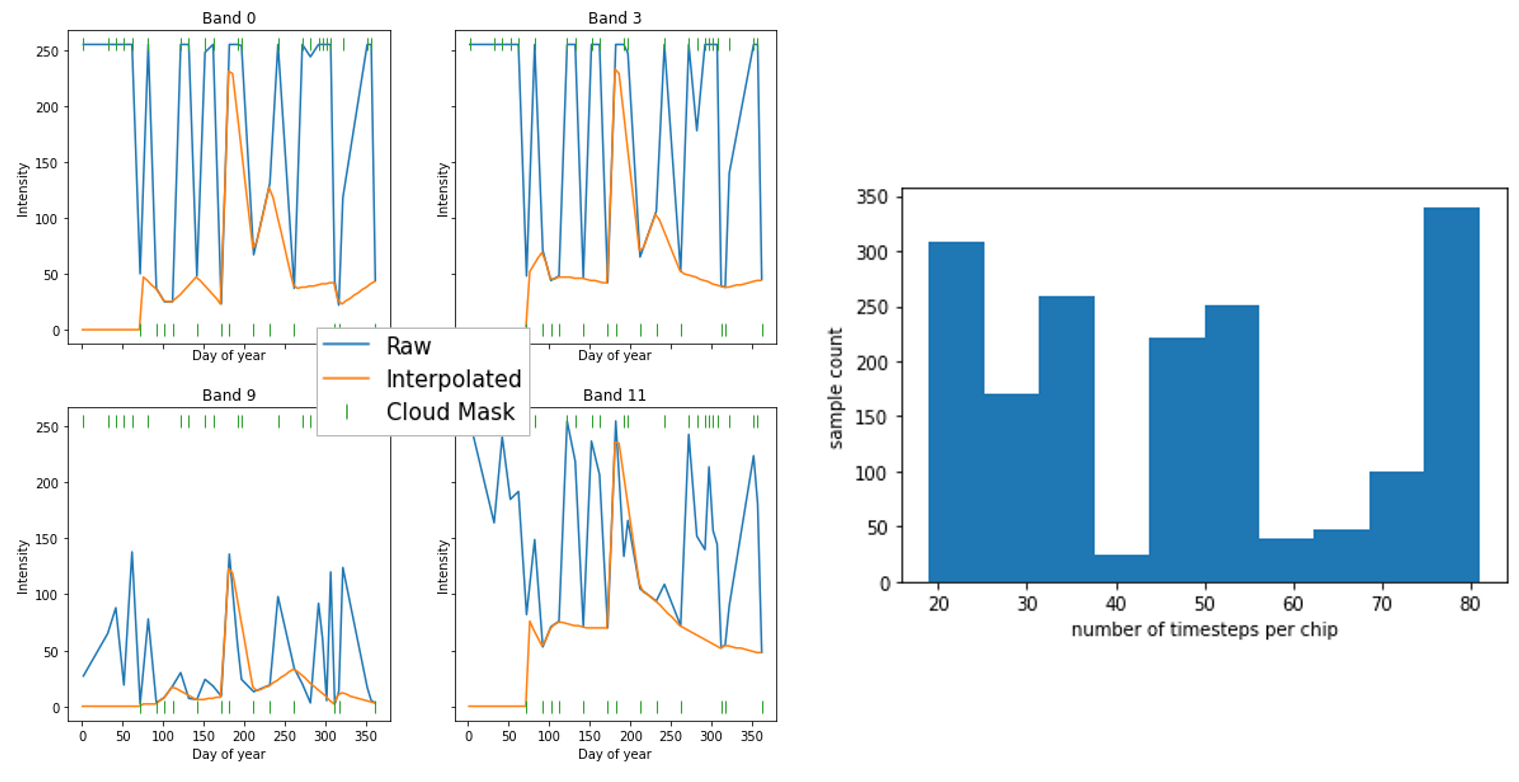}
  \caption{\textit{Left}: Spectral values for selected spectral bands over the course of the year, taking into account cloud presence. \textit{Right}: The gridded spatial subpatches (chips) have a minimum of 20 and maximum of 80 time steps.}\label{fig:spectra_over_time_and_timesteps}
\end{figure}

As an improvement over both spatial-spectral and spectral-temporal, the spatial-spectral-temporal \cite{crop_type_africa} model jointly optimizes over all available input features, allowing the model to incorporate neighboring pixel information, learn temporal growth characteristics, and account for spectral relationships.
However, this model comes at a cost of vastly increasing model size and complexity, requiring more training data and making it harder to successfully train.
Because the input requires a temporal component in addition to the spatial and spectral components, the satellite image time series input must be temporally interpolated on a regular grid to ensure uniform time steps across all input samples.
As shown on the left side of Figure~\ref{fig:spectra_over_time_and_timesteps}, each pixel of each spectral band (only 4 of the 12 bands are shown) must be separately interpolated across space and time to account for clouds.
The right side of Figure~\ref{fig:spectra_over_time_and_timesteps} shows the variability of time steps across all spatial-spectral-temporal samples, which requires introducing another hand-tuned parameter of minimum allowable number of time steps.
For these reasons, the spatial-spectral-temporal model will not be shown in any evaluations as it had poor training performance on our data.
This remains a promising approach but requires more work to yield valuable results.

\subsection{Evaluation in France}
\begin{figure}[H]
  \centering
  \includegraphics[width=8cm]{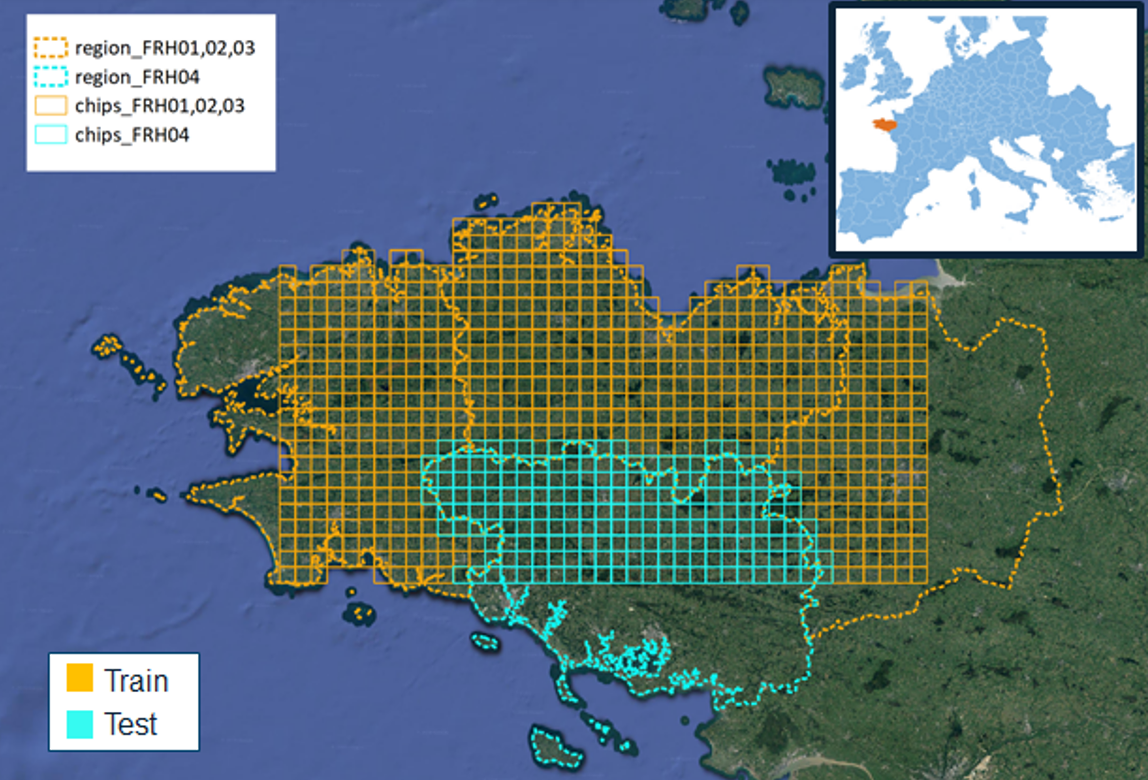}
  \caption{Three subregions are used for training, and one held out subregion is used for testing.}\label{fig:brittany_train_test}
\end{figure}

The first experiment was constrained to the region of Brittany, France, following the spatial area and crop classes used in \cite{breizhcrops}.
The train and test regions are shown in Figure~\ref{fig:brittany_train_test}, also following the subregion splits used in \cite{breizhcrops}. 

\begin{figure}[H]
  \centering
  \includegraphics[width=10cm]{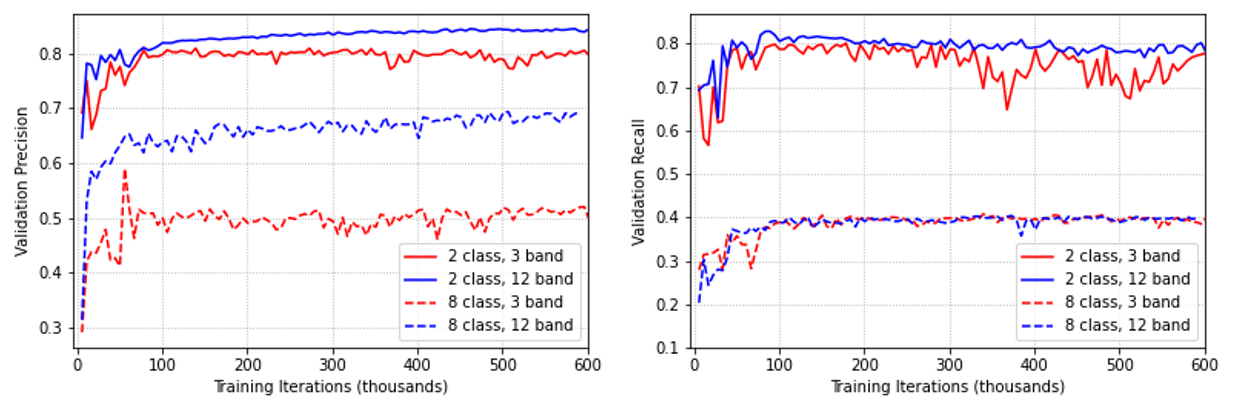}
  \caption{In both precision and recall, training with 12 bands and 2 classes performs the best.}\label{fig:france_training_plots}
\end{figure}

Four spatial-spectral models were trained to investigate spectral importance and how the number of classes affects model performance. The two class model consists of "wheat" and "background" only, while the eight class model includes additional crop types such as "barley" and "vegetables".
As shown in Figure~\ref{fig:france_training_plots}, the 2 class/12 band model showed the most promise.
For all future spatial-spectral experiments, we use this combination of classes (wheat, background) and number of spectral bands.
Figure~\ref{fig:brittany_train_test} shows examples of predictions for four random patches from this model.

\begin{figure}[H]
  \centering
  \includegraphics[width=12cm]{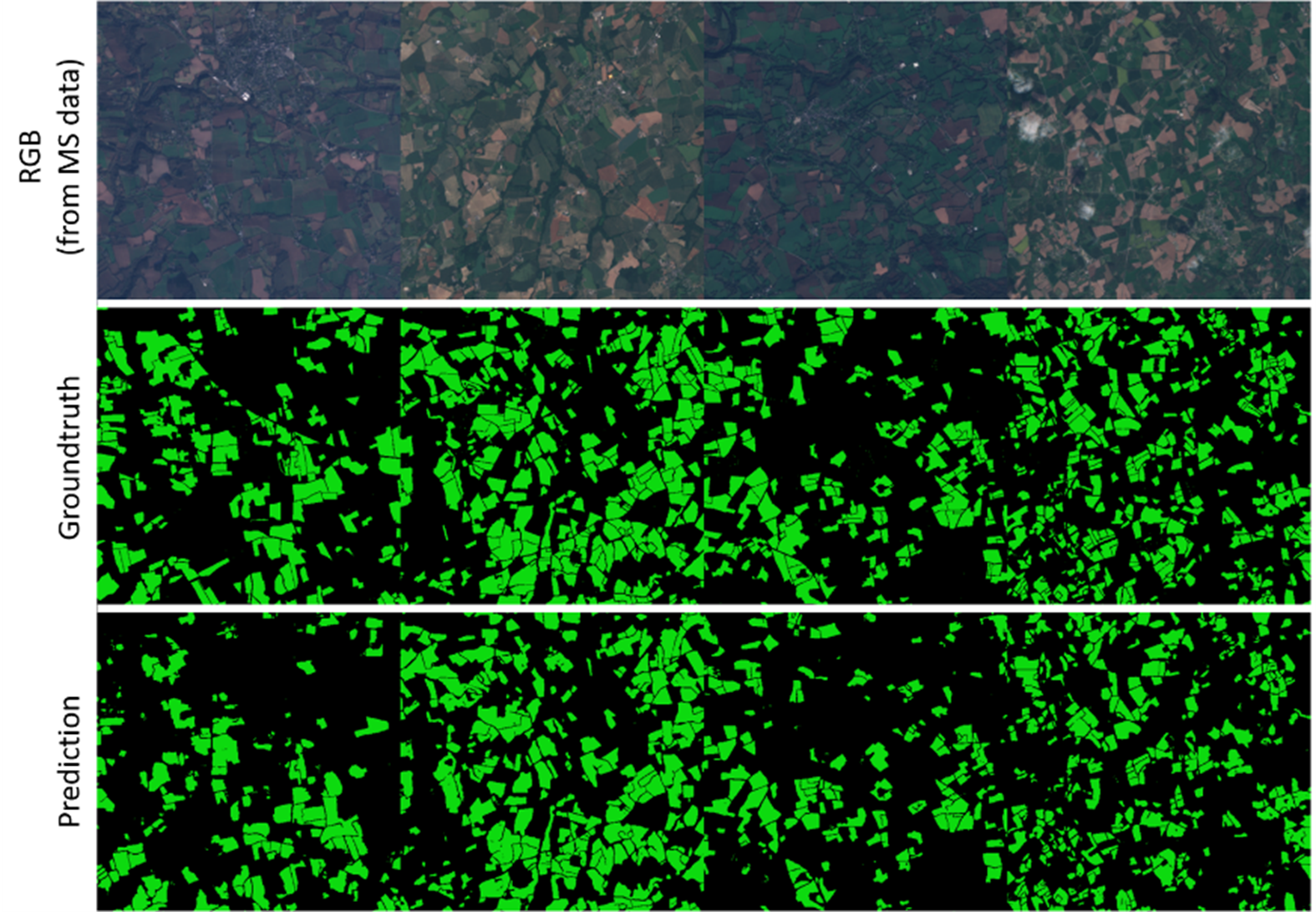}
  \caption{Predictions of wheat in various growth stages and regions closely follow ground truth.}\label{fig:spatial_spectral_prediction_chips}
\end{figure}

Following \cite{breizhcrops}, we use an LSTM spectral-temporal model and train on the same region, train/test splits, and classes.
Here we use the metrics of precision, or the proportion of positive classifications actually correct, and recall, the proportion of all positives correctly classified.
As shown in Table~\ref{tab:france_model_comparison}, the spectral-temporal model outperforms the spatial-spectral model in both precision and recall.

\begin{table}
  \caption{Comparison of spatial-spectral and spectral-temporal validation results.}
  \label{tab:france_model_comparison}
  \begin{center}
    \begin{tabular}{ |l | l | c | c |}
    \hline
      Model Type & Model Classes & Precision & Recall \\
    \hline
      Spatial-Spectral  & All classes & 0.69 & 0.40 \\
      Spatial-Spectral  & Wheat only  & 0.84 & 0.68 \\
      Spectral-Temporal & All classes & 0.83 & 0.83 \\
      Spectral-Temporal & Wheat only  & 0.98 & 0.94  \\
    \hline
    \end{tabular}
  \end{center}
\end{table}

\begin{figure}[H]
  \centering
  \includegraphics[width=8cm]{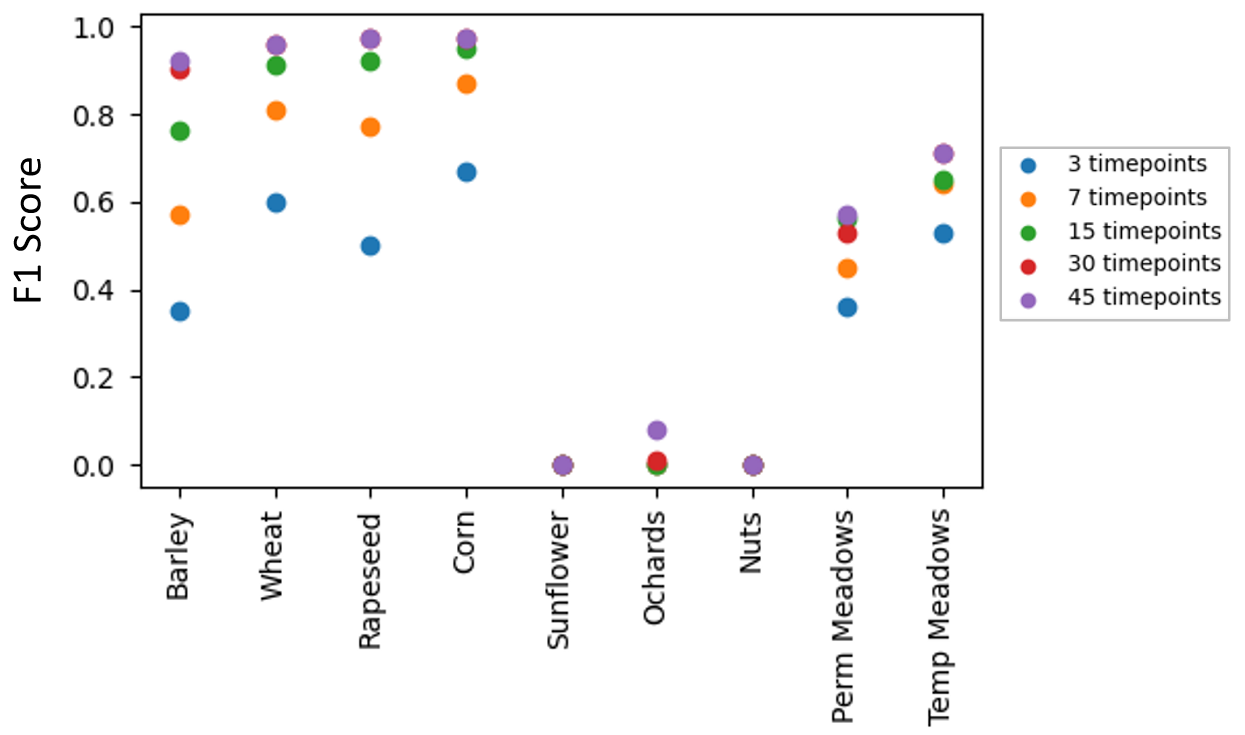}
  \caption{The spectral-temporal model maintains decent performance even with limited number of input time steps during training.}\label{fig:lstm_timepoints_study}
\end{figure}

Because the spectral-temporal model outperformed the spatial-spectral in this first set of experiments, it was important to understand its limitations before expanding its use to the northern India area of interest.
We study the importance of time period and number of time steps in the spectral-temporal model by limiting the time points during training and plotting the validation F1 score in Figure~\ref{fig:lstm_timepoints_study}.
The F1 score is the harmonic mean of precision and recall, allowing a single combined metric.
While the focus of this paper is the wheat crop, we use all classes in this experiment to understand how the model represents different types of vegetation in the temporal domain.
As shown in the plot, the spectral-temporal model can maintain comparable performance even when reducing temporal range by 33\% from 45 to 30 time points.

\subsection{Evaluation in India}

Now with an understanding of performance in data rich areas, we apply these trained models to northern India, which lacks any significant accurate parcel-level ground truth in the formats and scales required to understand model performance.
As a workaround, we evaluate using several approximate, pseudo-ground truth approaches.

\subsubsection{MapSPAM}

\begin{figure}[H]
  \centering
  \includegraphics[width=12cm]{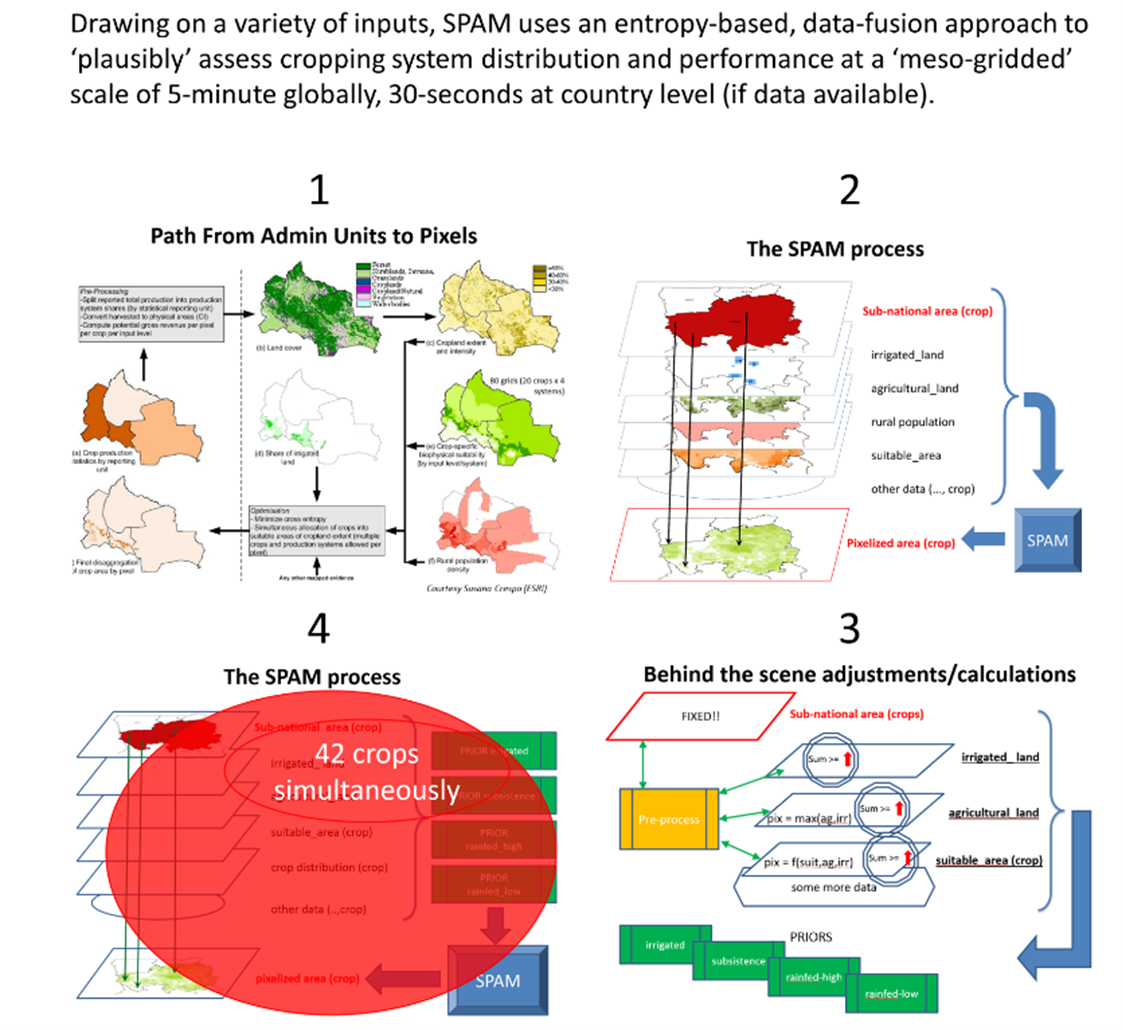}
  \caption{MapSPAM system overview. Text and graphic from https://www.mapspam.info/}\label{fig:mapspam_overview}
\end{figure}

MapSPAM 2010 \cite{mapspam} is a global crop type estimation method and data source developed using a variety of information sources.
It provides an approximate 10x10 km per pixel estimate, with each pixel value representing the percentage of each crop type in the pixel area.
The latest global model was produced for 2010 data, as the process is manually intensive and relies on collecting vast amounts of disparate data.
An overview is shown in Figure~\ref{fig:mapspam_overview}.
The potentially stale information and relatively low spatial resolution qualify this as a pseudo-ground truth, but we use this and several other approaches to quantify crop type mapping performance in northern India.

\begin{figure}[H]
  \centering
  \includegraphics[width=8cm]{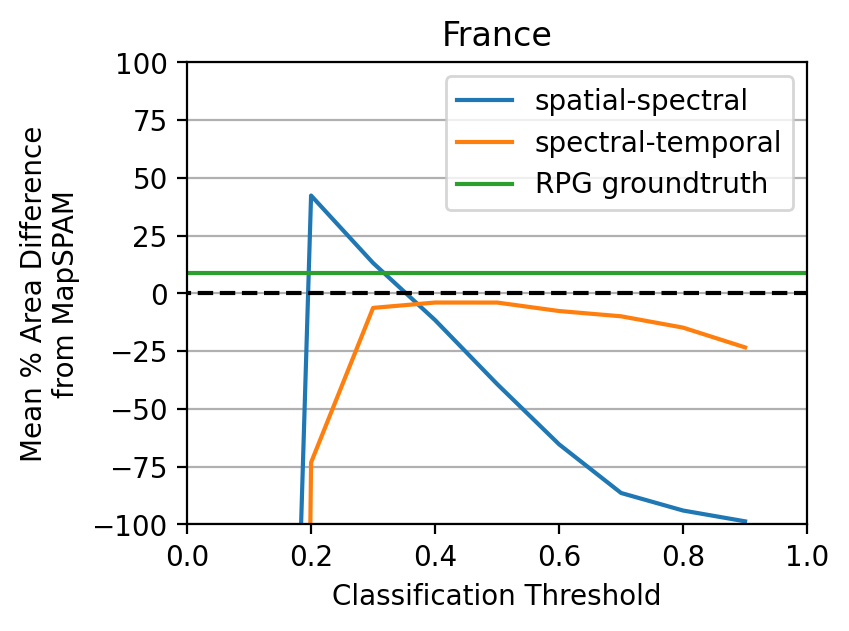}
  \caption{The spectral-temporal model is more accurate in total wheat area estimate than MapSPAM for a wide range of classification thresholds. }\label{fig:mapspam_vs_rpg}
\end{figure}

To better understand MapSPAM’s limitations, we first evaluate the spatial-spectral (SS), spectral-temporal (ST), and the actual RPG ground truth in France against MapSPAM’s predictions.
As shown in Figure~\ref{fig:mapspam_vs_rpg}, the real ground truth is about 10\% off from MapSPAM’s estimation in the French region evaluated. In Figure~\ref{fig:mapspam_vs_rpg}, we vary the threshold at which we classify a prediction as wheat or not wheat.
The SS model has a sharp peak around a classification threshold of 0.2, and otherwise vastly underestimates the area of wheat growth compared to MapSPAM.
The ST model’s estimation is much smoother across a wide range of prediction confidence thresholds and remains close to the actual, indicating a more stable feature representation within the model.

\begin{figure}[H]
  \centering
  \includegraphics[width=8cm]{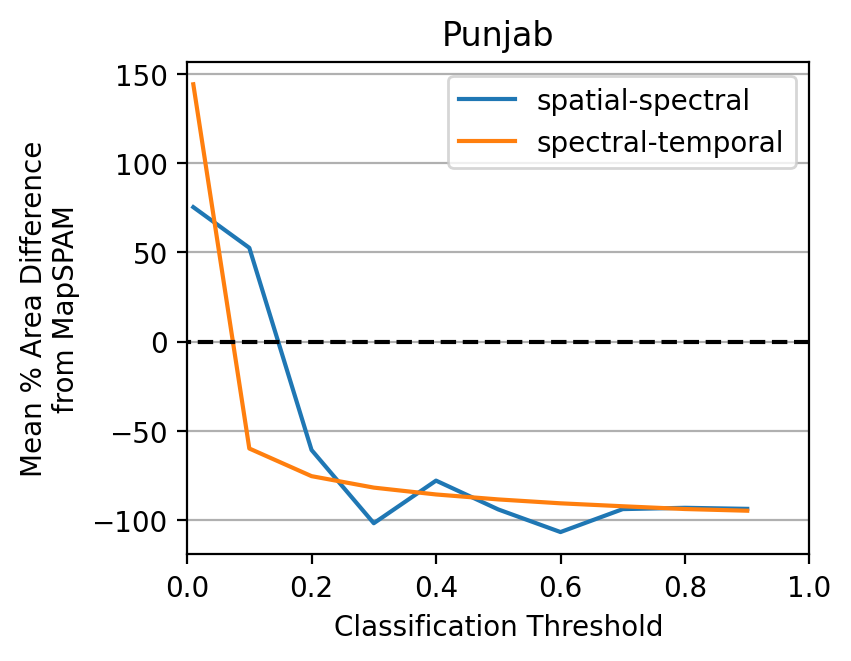}
  \caption{The MapSPAM area prediction intersects the 0\% area difference dashed line at low classification thresholds. This indicates a performance drop relative to the performance on the training data in Fig~\ref{fig:mapspam_vs_rpg}. }
  \label{fig:mapspam_vs_punjab}
\end{figure}

However, when applied to the northern India state of Punjab, both models accurately estimate the MapSPAM area at relatively low classification thresholds and have a large range of predictions sensitive to the classification threshold, indicating a performance drop across regions.
In the Punjab case, the SS model outperforms ST as evidenced from a higher confidence threshold at the MapSPAM’s estimation amount, as shown in Figure~\ref{fig:mapspam_vs_punjab}.

\subsubsection{NDVI as Upper Bound}

\begin{figure}[H]
  \centering
  \includegraphics[width=8cm]{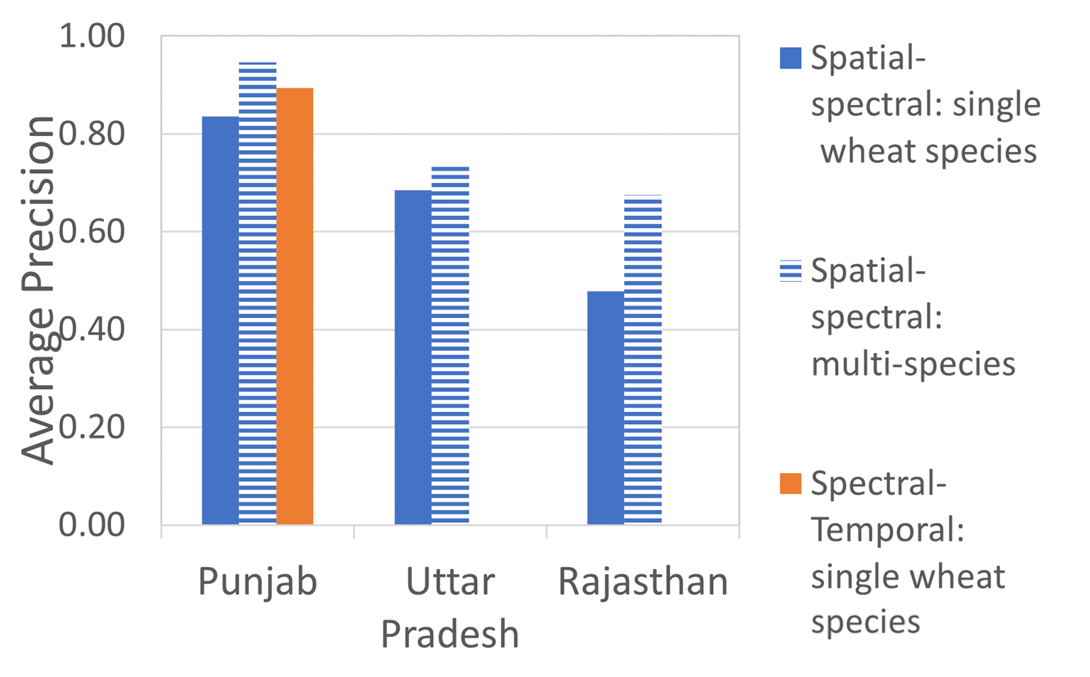}
  \caption{Using the NDVI map as pseudo-groundtruth, the model that treats all wheat species as a single class performs best across all Indian states studied.}\label{fig:ndvi_india_precision}
\end{figure}

Another pseudo-ground truth source is Normalized Difference Vegetation Index (NDVI) as an upper bound.
NDVI is defined as (NIR – Red) / (NIR + Red). A high value indicates strong vegetation presence (via spectral response of chlorophyll) in the pixel.
We filter NDVI to the range of values expected of wheat corresponding to a certain growth stage and time period.
These rough NDVI predictions should cover all of the wheat pixels, so our model should not estimate more area than the NDVI prediction.
We compare SS and ST models trained with a single wheat species and multiple wheat species (considered together as a single class) in Figure~\ref{fig:ndvi_india_precision} using the precision metric averaged across all images.
Precision was chosen since it compares only the model wheat predictions and does not consider recall as there are many more non-wheat vegetation pixels present in a region.

\begin{figure}[H]
  \centering
  \includegraphics[width=10cm]{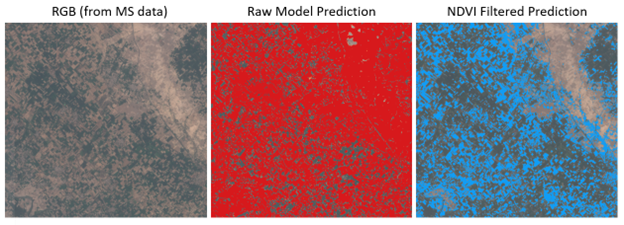}
  \caption{Qualitatively, the NDVI layer can help filter wheat prediction in obviously non-vegetation areas, as shown by barren land (light brown) in the imagery.}\label{fig:ndvi_filtered_prediction}
\end{figure}

Given the model behavior of predicting wheat in a few non-vegetation areas, we also consider using NDVI to filter the model predictions by removing any predictions not in the NDVI filtered mask.
We show one of the worst prediction areas from India in Figure~\ref{fig:ndvi_filtered_prediction}, and qualitatively demonstrate that using NDVI to filter some non-vegetation areas is a useful technique. 

\subsubsection{State-level Area}

\begin{figure}[H]
  \centering
  \includegraphics[width=8cm]{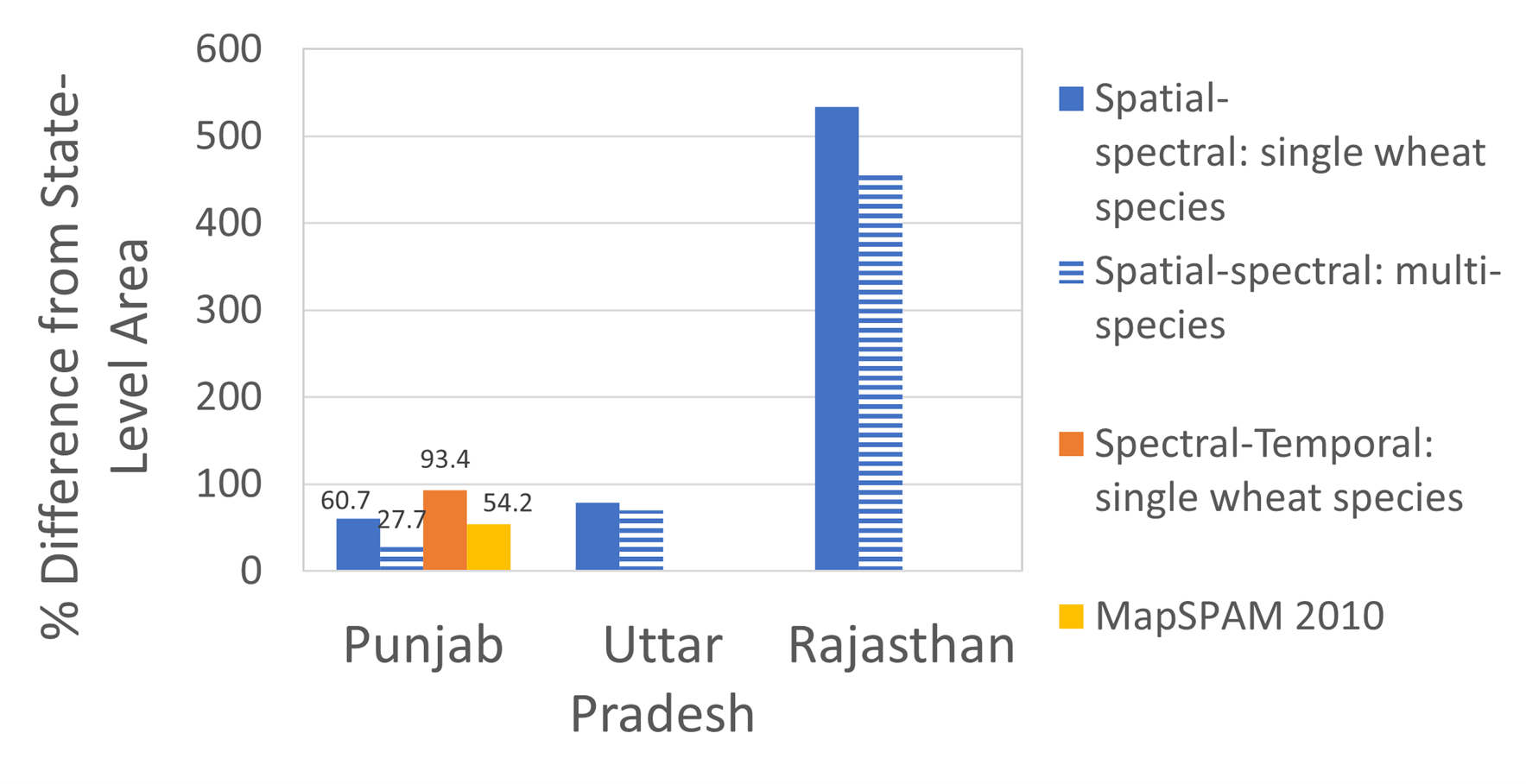}
  \caption{Qualitatively, the NDVI layer can help filter wheat prediction in obviously non-vegetation areas.}\label{fig:india_state_level_area_comparison}
\end{figure}

Finally, as a third pseudo-ground truth source we use state-level reports from the Indian Directorate of Wheat Development from 2019.
We process the wheat growing period of 2018-2019 and create final predictions for the same single species and multi-species SS and ST models and show results in Figure~\ref{fig:india_state_level_area_comparison}.
Overall, Punjab has the smallest difference from the state-level reports by area, as it is the most similar to the training regions in France. This similarity will be further examined in a following section.
We also include the MapSPAM 2010 predictions for Punjab, and they are off by about 50\% from the state reports.
It is important to consider error in the state-level reports as not all of the grown wheat is directly reported, but this error is much harder to quantify.

\subsection{Need for Domain Adaptation}

As shown in the previous pseudo-ground truth approaches, the models typically perform best on Indian states closest to the France training region, although they too suffer from domain differences.

\begin{figure}[H]
  \centering
  \includegraphics[width=10cm]{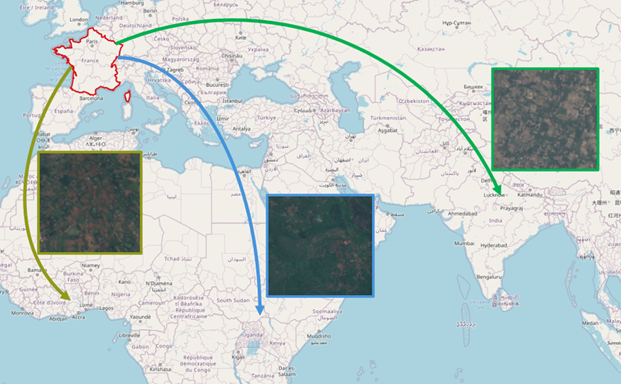}
  \caption{Agricultural features in data-rich areas such as France must be adapted to work in other domains across the world, with Sentinel-2 imagery from each location shown.}\label{fig:domain_differences_notional_map}
\end{figure}

In order to create an operational food security model, crop classification models need to work with different terrains, climates, and agricultural practices for broad applicability.
As previously explained, many regions of interest have little to no ground truth with which to develop data-intensive approaches such as neural networks.
To this end, Unsupervised Domain Adaptation (UDA) may be a promising approach to multi-domain crop classification.
UDA entails automated knowledge transfer from label-rich source domain to target domain with no labels, with a notional example shown in Figure~\ref{fig:domain_differences_notional_map}.

\begin{figure}[H]
  \centering
  \includegraphics[width=8cm]{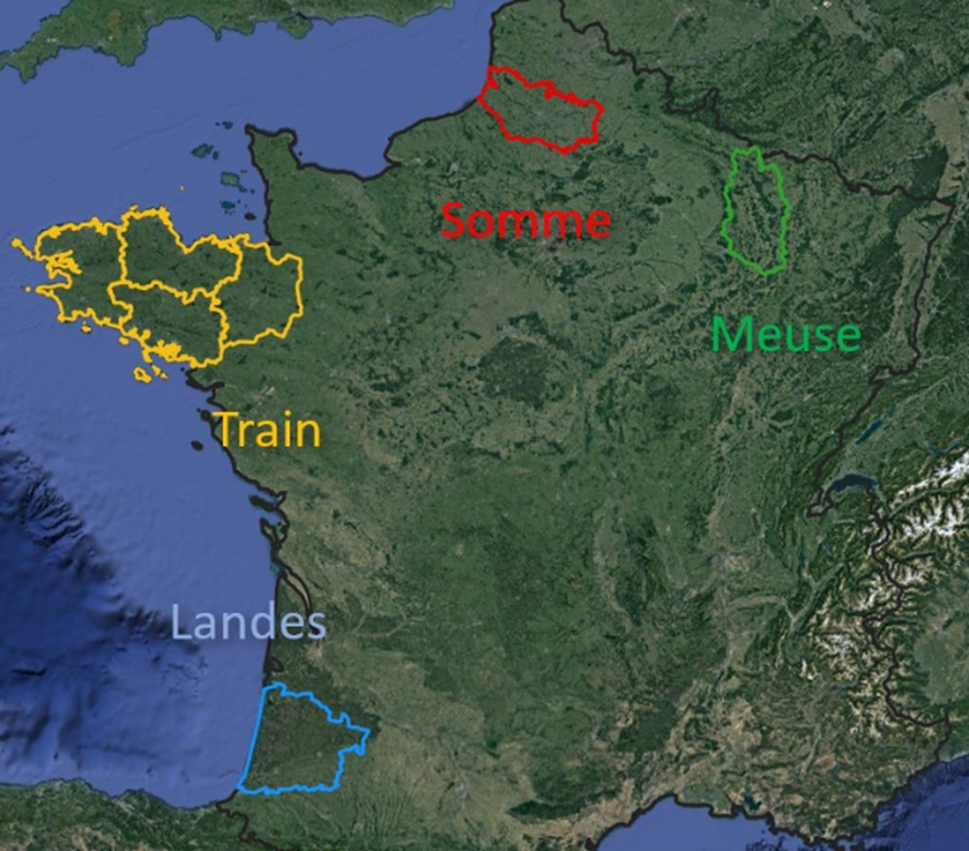}
  \caption{Training region of Brittany and testing regions of Somme, Landes, and Meuse.}\label{fig:three_france_regions_uda}
\end{figure}

To quantify and investigate the regional differences in wheat classification, we first compare sub regions in France due to its high fidelity ground truth.
We train an SS model in Brittany, and test on other areas in France to show drop in performance, as shown in Figure~\ref{fig:three_france_regions_uda}. 

\begin{figure}[h]
  \centering
  \begin{tabular}{ll}
  \includegraphics[scale=0.6]{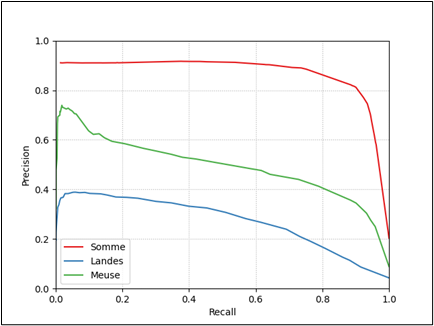}
  &
  \includegraphics[scale=0.5]{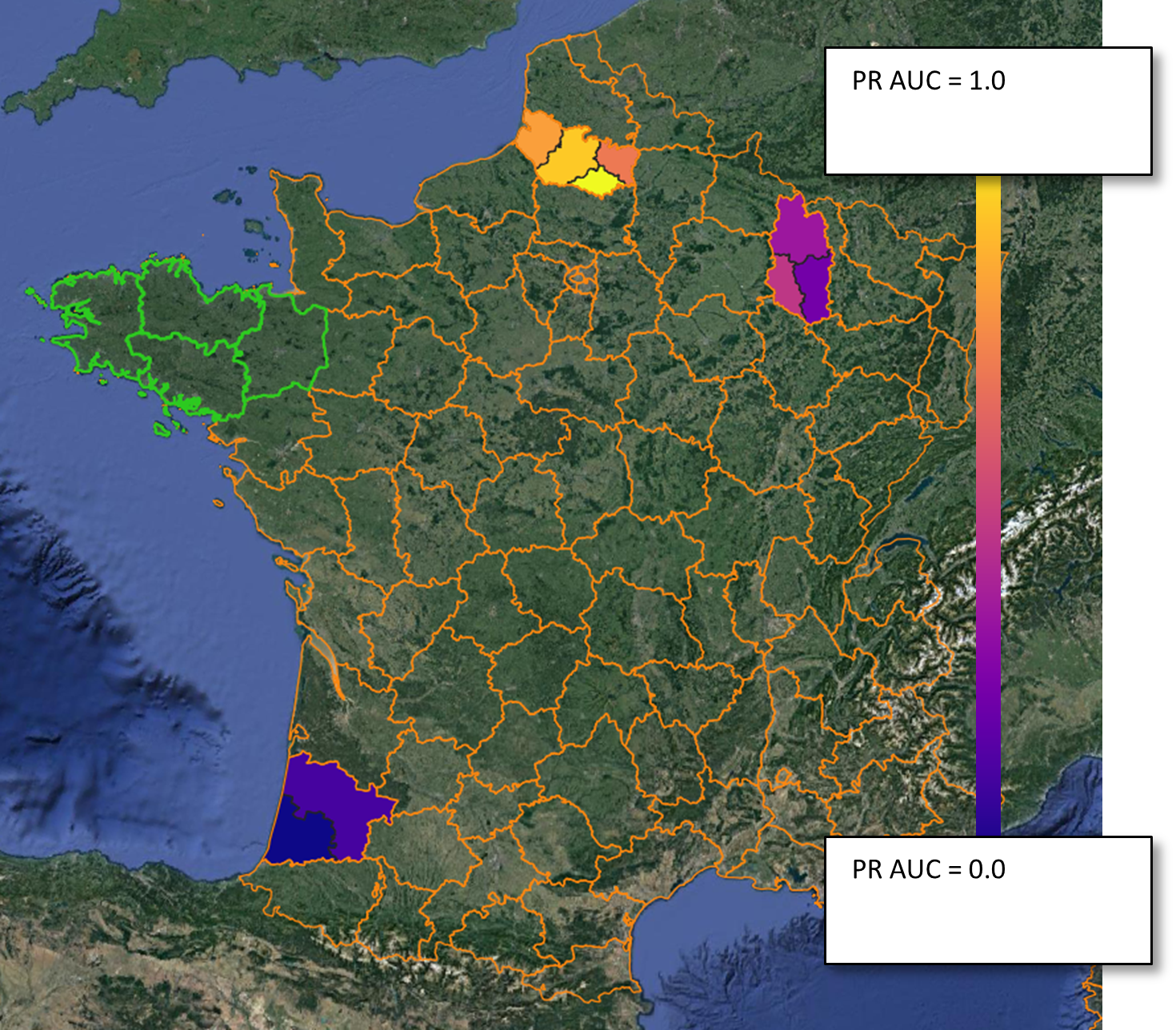}
  \end{tabular}
  \caption{Left: Results showing performance drop when testing in different regions from training. 
  Right: Map representation of precision-recall area under curve (PR AUC).}
  \label{fig:three_france_regions_results}
\end{figure}

We show that testing regions with a similar latitude to the training regions have less performance drop in Figure~\ref{fig:three_france_regions_results}. This result motivates a deeper dive into domain differences.

\subsection{Understanding Domain Differences}

To better understand these domain differences we try visualizing the learned features using a dimensionality reduction method called Uniform Manifold Approximation and Projection (UMAP) \cite{umap}.
UMAP is similar to a common dimemsionality reduction method called t-SNE in that it projects high dimensional data into a low dimensional embedding, but it also preserves both local and global distances between individual samples, making it a useful tool for visualizing high dimensional data on a 2D surface.

We compare features extracted from 400 samples (105M pixels) from the France training region and three Indian states of Punjab, Rajasthan, and Uttar Pradesh during their respective wheat growing seasons.

\begin{figure}[H]
  \centering
  \includegraphics[width=12cm]{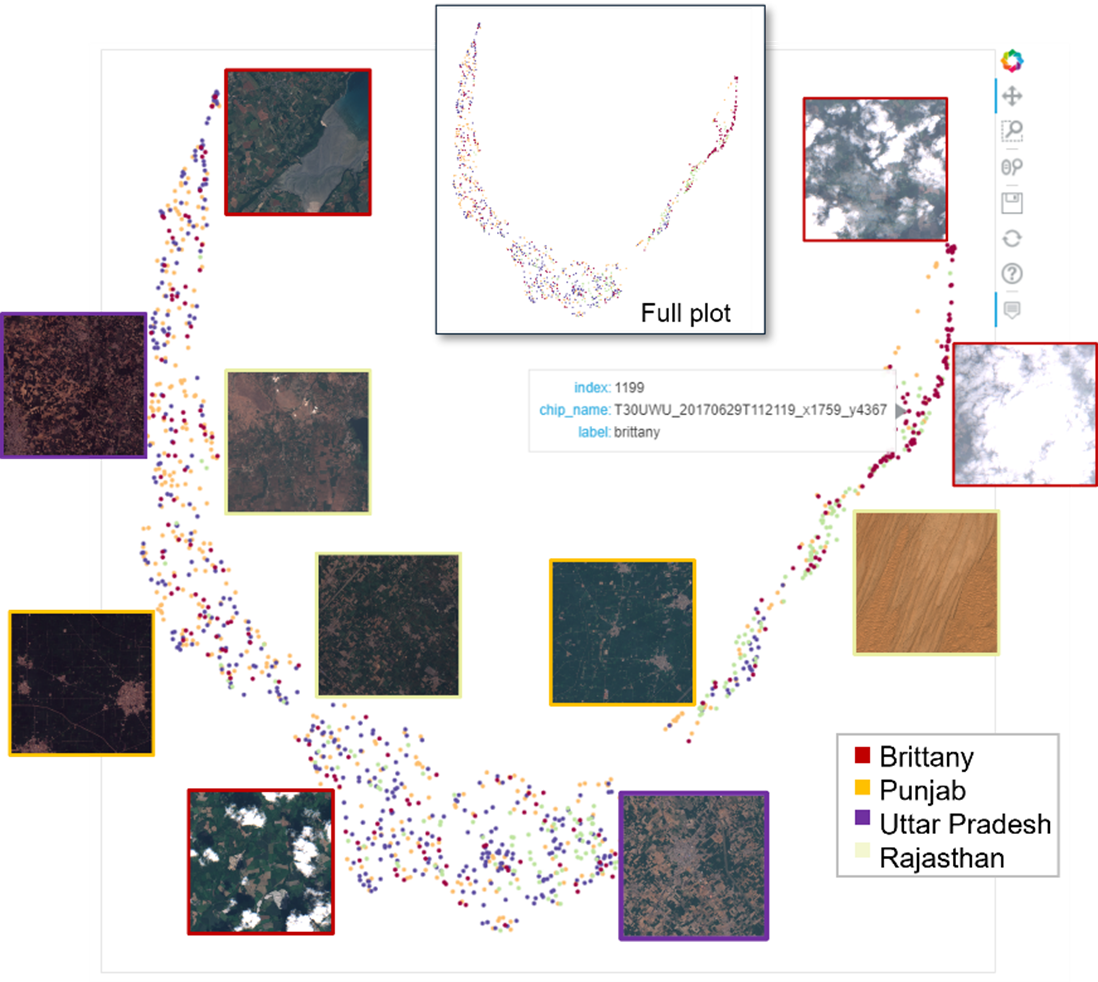}
  \caption{Learned features extracted from our CNN model cluster on a smooth spectrum, showing some regional differences.}\label{fig:umap_clusters}
\end{figure}

As shown in Figure~\ref{fig:umap_clusters}, there is a relatively smooth spectrum of samples, but some points group together for each region.
Clouds and other atmospheric effects were not directly taken into account, so some clusters may be due to cloud presence.

\begin{figure}[H]
  \centering
  \includegraphics[width=12cm]{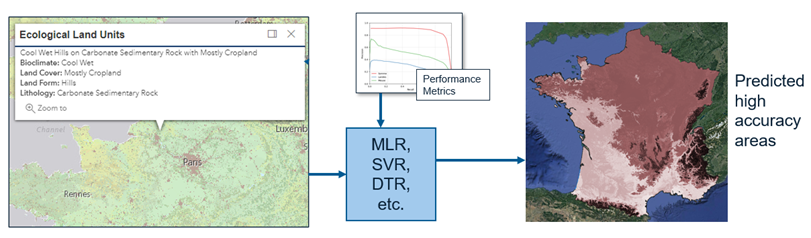}
  \caption{Notional graphic of using USGS ELUs to predict performance in other regions of interest.}\label{fig:elu_regression_notional_graphic}
\end{figure}

We also consider USGS Environmental Land Units (ELUs) \cite{usgs_elu}, which is a map of ecophysiographic stratification based on bioclimate, landcover, landform, and lithology.
The ELU data can explain some domain differences, but to quantify how these affect current model performance, we map F1 crop classification performance metric in different regions to ELU input variables.
A notional example is shown in Figure~\ref{fig:elu_regression_notional_graphic}, with the application of determining ELU variable importance for predicting future model performance in new areas.
We also plan to use a regression model to inform domain generalization design choices in future work.

\subsection{Generalization vs Adaptation}

As the ideal application of this system is to unique and disparate global regions, it is important for a crop identification model to accurately map wheat and other crops across many different domains.
There are two main perspectives for achieving robust models in this context: domain generalization (DG) and domain adaptation (DA).
The aim of DG is to have a single model that can be used in a variety of domains and achieve suitable performance in all.
As a contrast, DA aims to transfer knowledge from one domain to another, meaning a new DA training is performed for each separate domain. 

A common approach for pretraining a DG model is called contrastive learning \cite{contrastive_learning}, which is a self-supervised visual representation learning method that learns invariance by predicting the features of a transformed image.
As it is self-supervised, it requires no labels, making it effective at training on large scale datasets.
Contrastive learning pre-training outperforms supervised pre-training on many downstream tasks and datasets, proving a suitable choice for DG tasks.

As previously explained, UDA can provide automated knowledge transfer across a single source and multiple target domains without the need for large labelled datasets.
We studied two UDA approaches using different regions within the France RPG dataset.
The UDA setting is created by withholding the labels in training for the target dataset, and only using the labels in the final evaluation.

\begin{figure}[H]
  \centering
  \includegraphics[width=10cm]{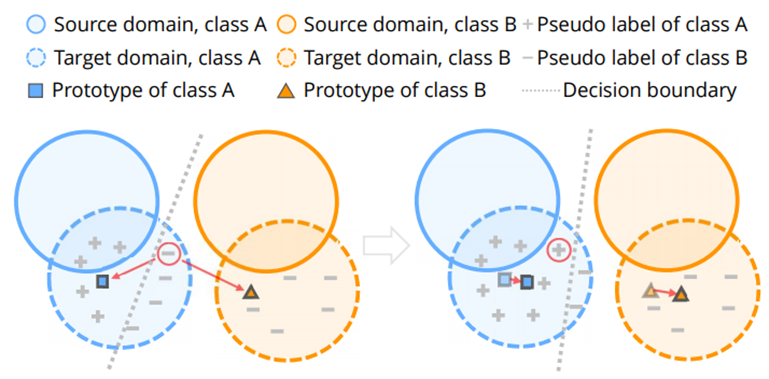}
  \caption{ProDA can use prototypes of classes in source and target data sets to align domains in feature-space.}\label{fig:proda_overview}
\end{figure}

A first approach is called ProDA \cite{proda}. ProDA defines classes in the target dataset using pseudo labels, created by iteratively minimizing the distance from a sample point and its prototype (centroid of feature clusters).
By using the distance from the prototype for each class, the pseudo-labels are corrected online during training and account for noisy outliers typically present in UDA datasets.
An overview is shown in Figure~\ref{fig:proda_overview}.

\begin{figure}[H]
  \centering
  \includegraphics[width=12cm]{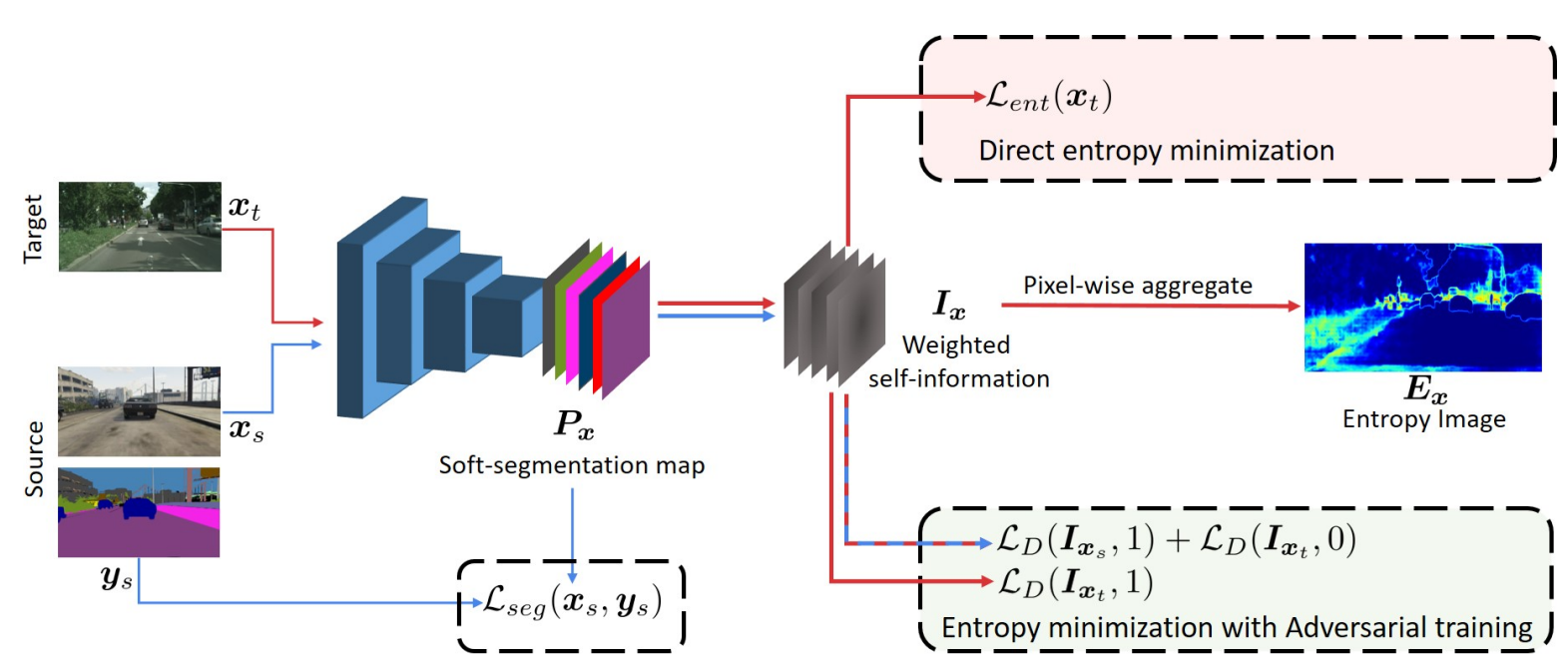}
  \caption{ADVENT attempts to align source and target domains in feature-space via entropy minimzation, allowing for very little additional computational overhead.}\label{fig:advent_overview}
\end{figure}

A second approach is called ADVENT \cite{advent}, which combines the well-practiced adversarial training regime to maximize domain overlap with a simple entropy minimization on the target predictions.
Figure~\ref{fig:advent_overview} shows an overview of the approach, where the joint loss combines a soft segmentation loss of the predicted target image and performs either a direct entropy minimization, or an entropy minimization via adversarial training.
The latter option attempts to align the entropy distributions of the source and target datasets computed on the self-information maps.
Overall, ADVENT is a useful approach as it is relatively simple in reducing entropy of pixelwise predictions across train and test domains, and adds no significant overhead to semantic segmentation training.

The area of domain adaptation/generalization for crop classification remains challenging.
While these approaches show promise in theory, both are active areas of research in applying to crop classification, and have yet to yield significant improvements over baseline networks in our effort. 

\section{Crop Models}

\subsection{Simulation Details}
While remote sensing and machine learning are appropriate techniques for identifying individual wheat fields, they do not provide the fidelity or the predictive capability that are necessary for quantifying wheat yields. For this, we turn to physics-based crop growth models, of which there are several in use by academic and industrial researchers. We considered three of the most popular models available (DSSAT, APSIM, and WOFOST) and chose WOFOST primarily due to the existence of a Python wrapper, PCSE, to make integration into our software pipeline easier~\cite{de2024pcse}.

The World Food Studies (WOFOST) crop model, from Wagneningen University \& Research, Netherlands, has been in use for over 25 years and is currently a key component of the MARS Crop Forecasting System for Europe~\cite{van1989wofost}. It provides potential yields, which are limited only by environmental conditions and plant characteristics, as well as attainable yields, which are also limited by water and soil nutrition conditions. The model uses daily time series weather data, soil data, and crop parameters to provide crop yields in kg/ha.

We collected and mapped soil data from the Harmonized World Soil Database’s ~\cite{hwsd1,hwsd2} Texture field to one of WOFOST's three built-in soil classes. Our testing focused on the winter wheat crop. We chose winter wheat 104 as the wheat strain to use after testing different strains across the country. We identified October 1st as an optimal planting date by testing each potential day, 9/16 through 11/30, against our ground truth data in France. 

For validation efforts, PCSE provided a built-in API linked to the NASA POWER database (https://power.larc.nasa.gov). This provided good historical weather data for all parameters used by PCSE that we compared to ground truth data. However, for future climatic conditions, we employed MarkSim, which can provide weather projections for any year on a daily basis~\cite{jones2000marksim}. This allowed us to run models for future crop yield output. 

Although MarkSim provided most of the requisite data, it did not provide top of atmosphere radiation, dew point, wind speed, and average daily temperature. Top of atmosphere radiation was assumed to be relatively constant and was mapped to the previous year’s values. Ground level radiation was provided by MarkSim. Dew point, wind speed, and average temperature were calculated using a probability distribution based on historical NASA POWER weather data for each site.

We compared crop projections from PCSE to 20 years of harvest data from France, spanning 2000-2019.  The data were obtained from the French Ministry of Agriculture and Food (https://agreste.agriculture.gouv.fr.)

\subsection{Crop Yield Modeling Analysis Details}

We wanted to determine if WOFOST could be used to project crop production trends one or two decades into the future. Therefore, we sought to identify which weather patterns in France were associated with increased error from the WOFOST simulation. We computed simulation error by comparing the predicted crop yield output for each department in France, for every year from 2000 through 2019, to the true crop yield in France in those regions. The year 2016 was omitted from the analysis due to the severe loss in wheat yield that year~\cite{ben2018causes,noia2023extreme,VANDERVELDE2020139}. France weather data were collected via the NASA Power API.

With absolute percent error as the response variable, we built a longitudinal penalized functional regression model using the \texttt{lpfr} function from the \texttt{refund} package in R~\cite{refund}. The predictors included a fixed effect for time, which was coded as a categorical variable; a random intercept for geographic department or region; and up to three functional variables: weekly average max temperature; monthly total precipitation; and weekly average irradiance. These data were aggregated to their respective timescales to allow the functions to be somewhat smooth and aid in constructing the regression model. Another example of using functional data analysis for crop yield predictions can be found in~\cite{park2023crop}. 

\begin{figure}[H]
  \centering
  \includegraphics[width=12cm]{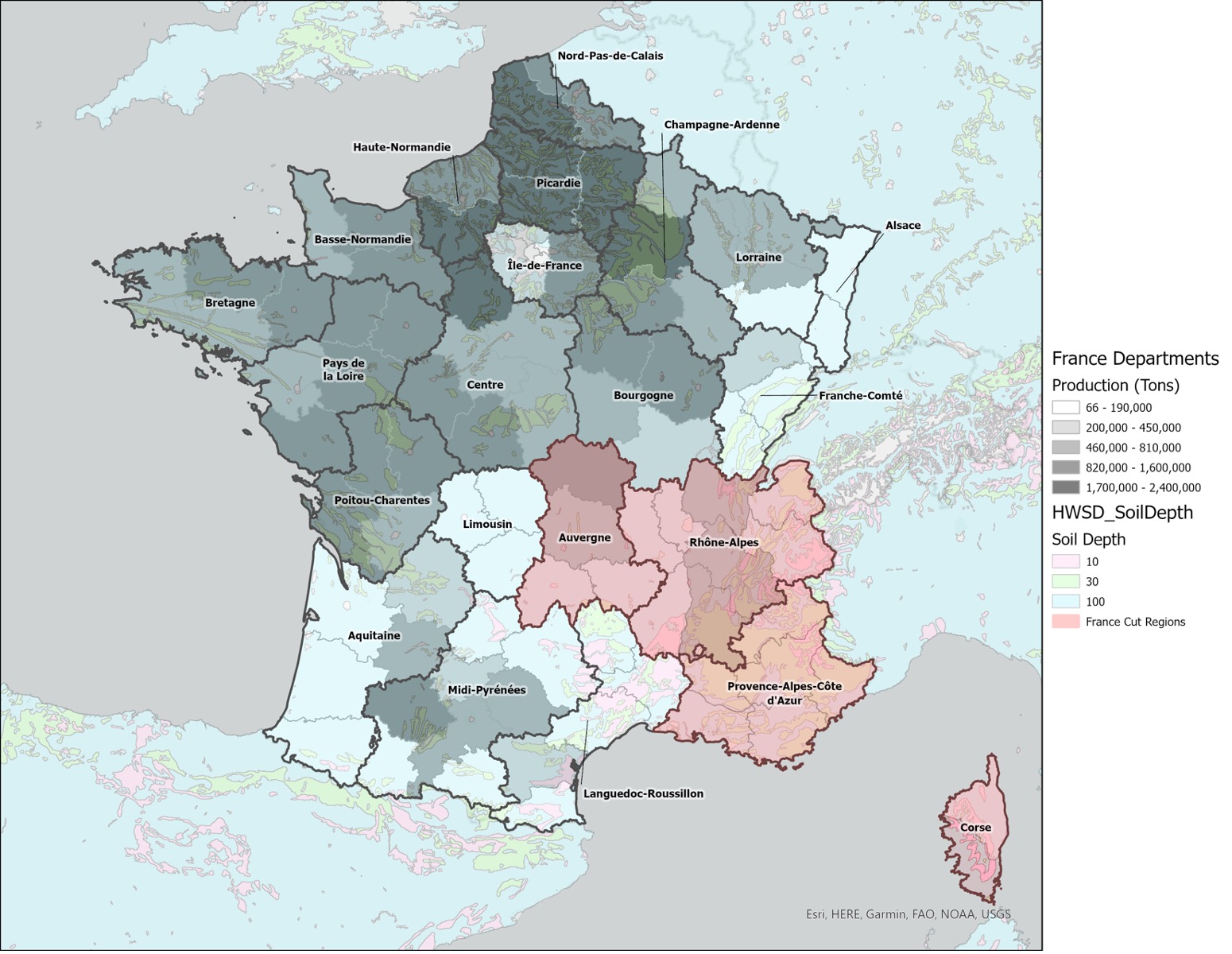}
  \caption{Regions of France overlaid with wheat production and soil depth}\label{fig:regions_soil}
\end{figure}

The analysis was done once at the department-level, using data aggregated and averaged by each of the 93 departments, and once at the level of the 13 regions. At the region-level, we removed three regions, which are all in the southeastern portion of France, from the analysis: Corse, Auvergne-Rh\^{o}ne-Alpes and Provence Alpes C\^{o}te d’Azur. Each of these regions are low producers of wheat, and they also tend to be regions of low soil depth, as shown in Figure~\ref{fig:regions_soil}. Therefore, since we focused on the bread-basket regions of France, we omitted these regions. 

The region-level analysis identified that the categorical variable of time was critical to the model, affirming that each year can be quite different in terms of weather and crop production. However, after including time in the model – and therefore adjusting for its effects – rain and irradiance were identified to be significant predictors in the model as well. Removing rain or irradiance from the model worsened model fit; additionally, temperature was not found to be a significant variable in the region-level analysis. The adjusted $R^2$ of .68 indicated that the model with time, rain, and irradiance explained 68\% of the variation in absolute percent error at the regional level. 

\begin{figure}[H]
  \centering
  \includegraphics[width=12cm]{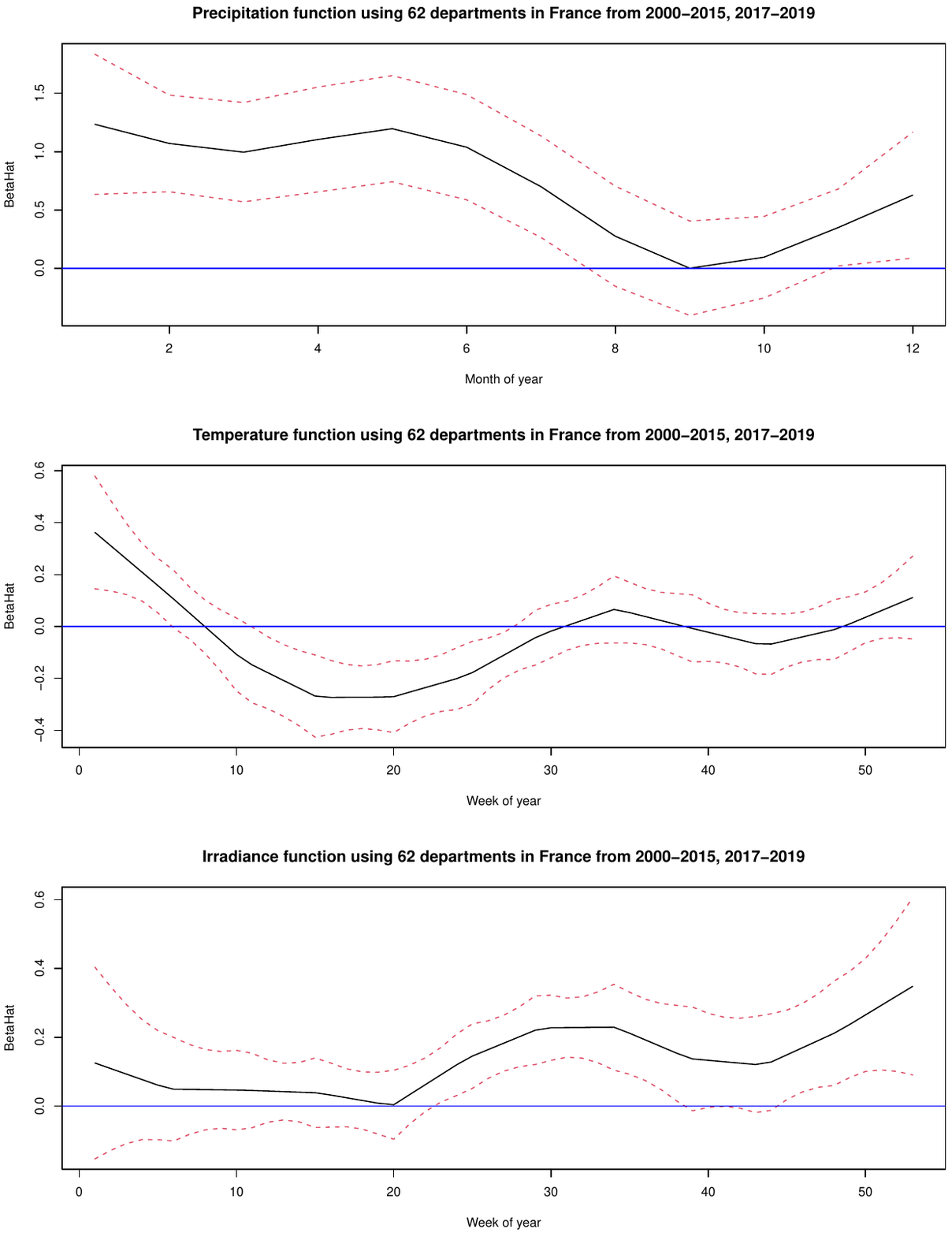}
  \caption{Coefficient plots for precipitation and irradiance using 62 departments in France}
  \label{fig:region-coefficient-plots}
\end{figure}

Additionally, coefficient plots (Figure~\ref{fig:region-coefficient-plots}) help explain when and how these weather factors may be associated with absolute percent error. An increase in rain is associated with more error throughout the year, whereas temperature is associated with error from Week 16 to Week 40, or from May to the end of September. The coefficient plots can be interpreted following the rules in~\cite{dziak2019scalar}. 

The analysis at the department-level required several additional steps. First, all departments within the low soil-depth regions above were similarly omitted from this analysis. Next, Haute-Garonne was removed since its average percent error across the 19 years was over 10000\%. We then built a model which included the remaining 73 departments and examined the model for departments which contained at least one outlier-year. This process of identifying outliers was conducted until we obtained a model without any notable outlier-years that hampered the model fit.

\begin{figure}[H]
  \centering
  \includegraphics[width=12cm]{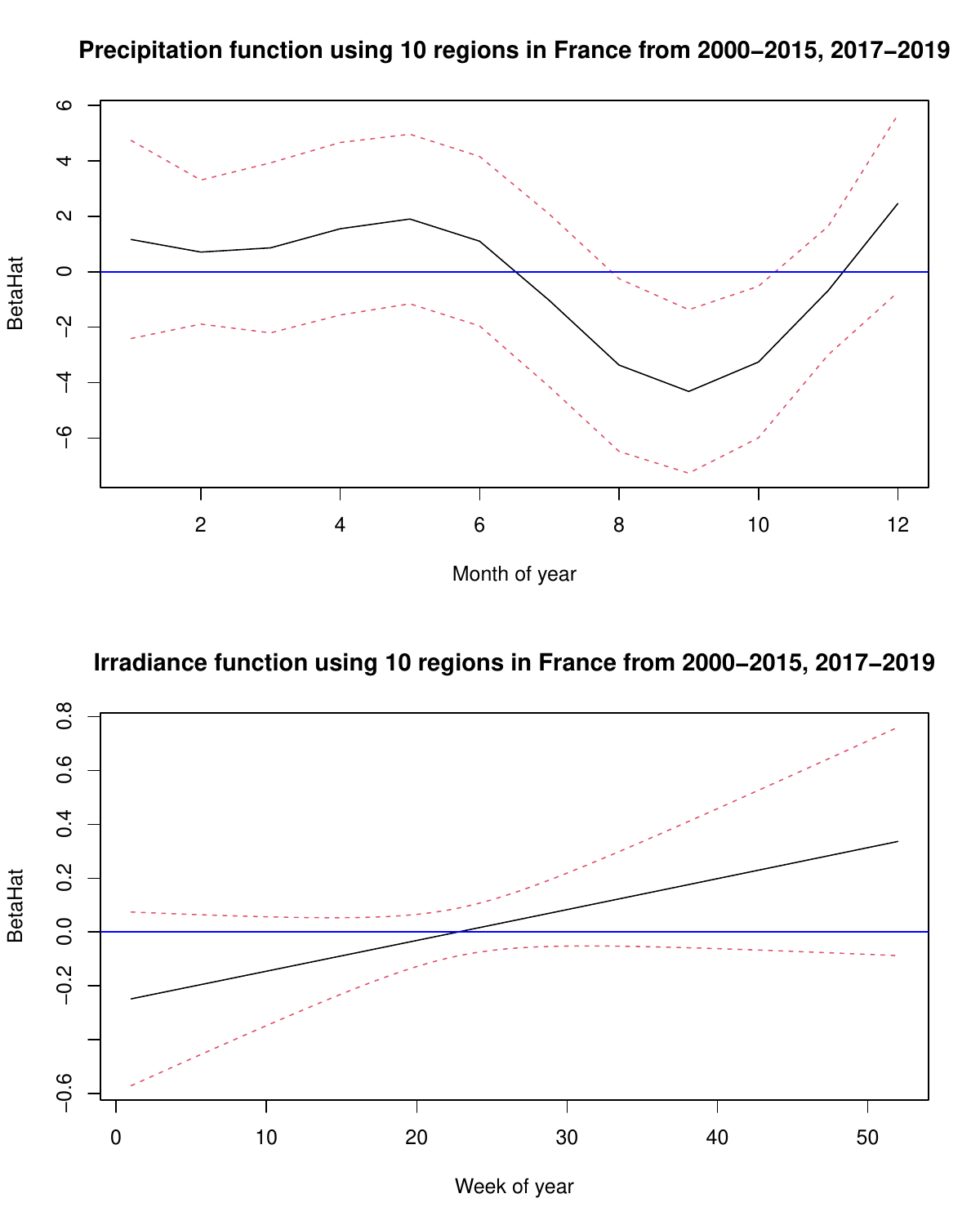}
  \caption{Coefficient plots for precipitation and irradiance using 10 regions in France}
  \label{fig:dept-coefficient-plots}
\end{figure}

\begin{figure}[H]
  \centering
  \includegraphics[width=12cm]{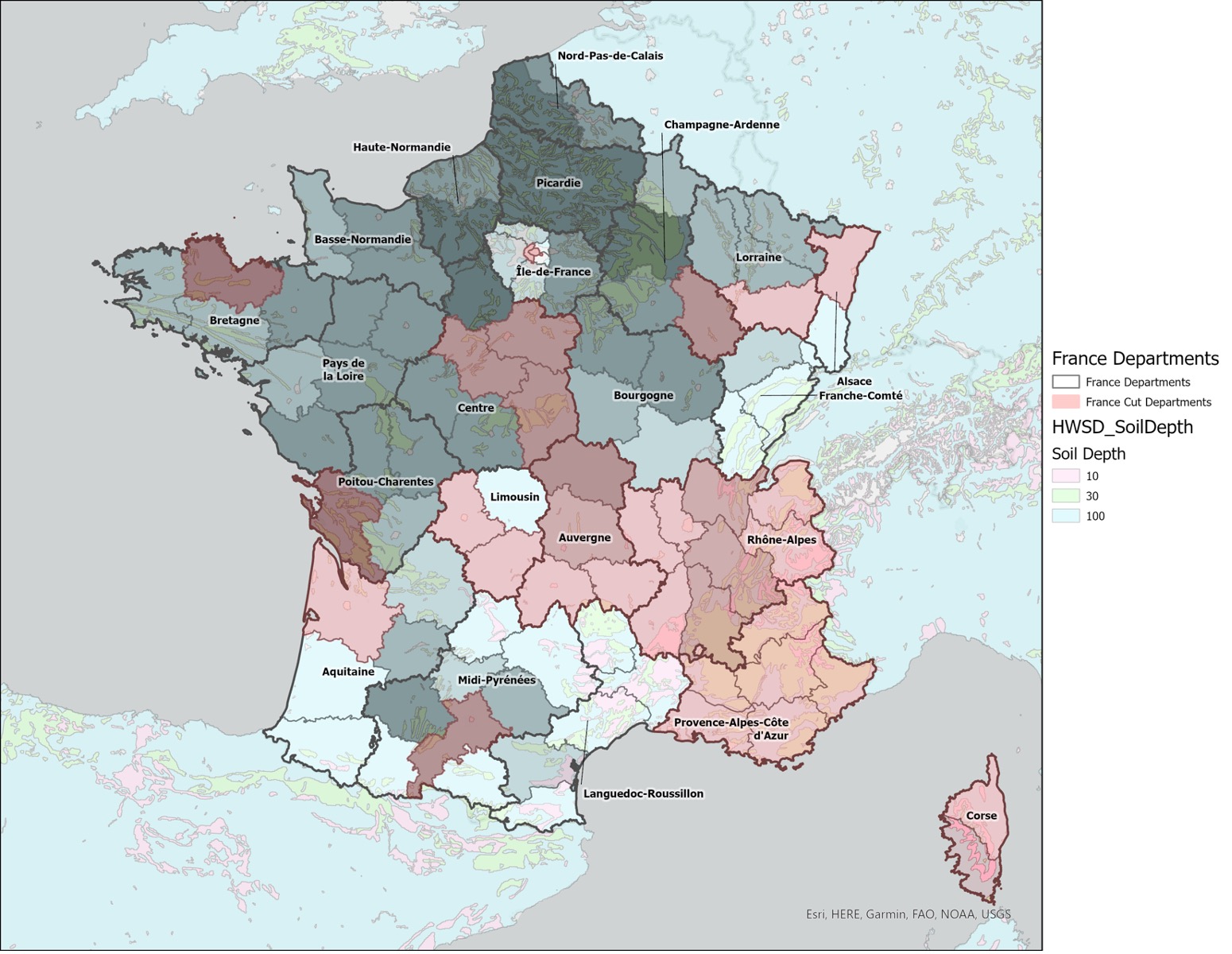}
  \caption{Regions of France overlaid with wheat production and soil depth}\label{fig:depts_soil}
\end{figure}

In total, 62 departments were included in the final department-level model, as shown in Figure~\ref{fig:depts_soil}. The conclusions here resemble those in the regional-level model: rain and irradiance are again included in this model, along with temperature. The adjusted $R^2$ value of .78 indicates that the model with time, rain, irradiance, and temperature explained 78\% of the variation in absolute percent error at the department-level. We can examine coefficient plots in Figure~\ref{fig:dept-coefficient-plots} to learn when and how the weather factors are associated with changes in error. An increase in rain is associated with more error throughout the year; temperature is associated with changes in error prior to Week 30 (August); and irradiance appears important all year round. 

By simulating crop yield output with PCSE and comparing crop yield errors to local weather data, we have built a roadmap for this type of crop production modeling. Given a reliable simulator, a representative crop strain, weather data, and a map of soil depth, one can identify the impact of changes in weather on crop production in different geographic regions. This kind of work can be used in anticipation of climate change – identifying weather patterns which hinder food production can allow for measures to be taken to prevent food shortages.

\section{The System Dynamics of India's Public Distribution System}

The dissemination of food to the vulnerable population is just as important for food security as predicting food production.  
This dissemination is done by a government program called Public Distribution System (PDS).  
The Public Distribution System was established in the middle of the 20th century by the government of India to deliver grain to its food insecure citizens \cite{bhatia1985food}.   
Over the following decades, the Indian government expanded and formalized the program to improve food security. Today, roughly two thirds of the country's population of more than 1.4 billion is entitled to a grain subsidy under the National Food Security Act (NFSA) \cite{puri2022india}.
Then in addition to the intention of the PDS to provide for the food insecure, the PDS also has a large impact on the food economy of India.

Hence, we model it to estimate food insecurity trends of India's citizens.  
The country of India is a collection of 28 states and eight union territories which have their own governments. 
These states and territories are further partitioned into administrative divisions called districts.  
Because states can be large and demographically diverse, we model the PDS at the local level of districts, which should be more helpful government officials.  

The model may also be able to suggest prescriptive policies that a government can apply. 
For this purpose, we've developed system dynamics models to identify potential levers for policymakers and to show effects of these decisions on food security.
Our approach is to build individual modules to model different aspects of food distribution. 
This allows modelers to swap out modules as new complexities are identified, or to enable new scenarios, such as modeling new geographic regions or expanding the model to include livestock or other crops. 

\begin{figure}[H]
\hspace{-1.7cm}
\includegraphics[width=15cm]{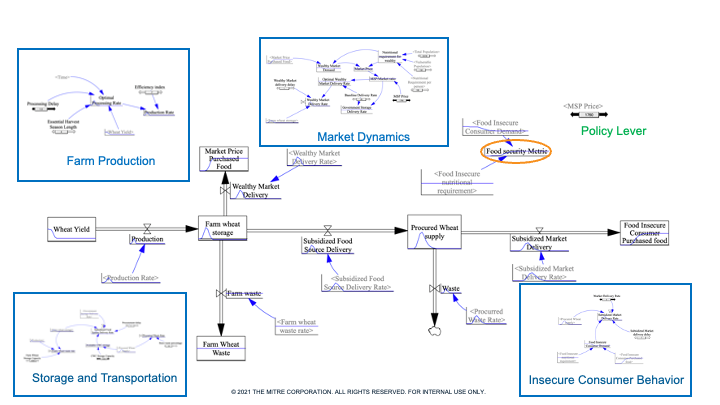}
\caption{Modules for the overall system dynamics model that describe the production and distribution of wheat in northern India}\label{fig:System_Dynamics_Modules}
\end{figure}

India's crop production and dissemination requires an understanding of the market dynamics as well as the dissemination policies of the government. 
The Indian federal government provides a Minimum Support Price, or MSP, to the farmers to encourage them to produce certain crops. 
Without the MSP, it is likely that many farmers would forgo growing staples such as wheat and focus on cash crops instead. 
Figure~\ref{fig:System_Dynamics_Modules} shows our four modules -- Farm Production, Market Dynamics, Storage and Transportation, and Food Insecure Consumer Behavior -- as well as the overall dissemination pipeline into which these modules feed. 
The MSP policy lever can be modified to assess how changing the MSP affects the availability of wheat for the food insecure.  

Building such a model has two main challenges.  
First, grain consumption trends inform how grain should be distributed within each state, but publicly available data on this is scarce at the district level.  
Therefore, we estimate food consumption trends using related data, which comes from a variety of data sources and varying spatial scales.  
Second, the transportation system of grain from farms to consumers is complex.
Grain produced in one state may need to be transported to a different one, which could depend on many factors including grain production in neighboring states, the weather, and current economic and infrastructural conditions.  
To make the problem more tractable, we focus only on the wheat crop and the state of Uttar Pradesh.  
Uttar Pradesh has a large, economically diverse population that produces and consumes wheat, which makes it a good test case for modeling the PDS for the rest of India \cite{pandey2012uttar}. 

Section~\ref{sect:sd_population} outlines how we estimate grain consumption data that is missing from public sources at the district level.  
This is done by aggregating information from multiple spatial scales: district, state, and regional level, as listed in Table~\ref{tab:description_of_data}.  
Section~\ref{sect:sd_model_pds} discusses our model for the flow of wheat through Uttar Pradesh, where we model only the essential features. 
A stock and flow diagram describes the within district behavior and a fully connected graph describes the transportation between districts.  
The final section, Section~\ref{sect:sd_results}, presents the results.  
Our model predicts the percent undernourished for the year 2019 using input data from previous years.  

\begin{table}[H]
\centering{
\caption{Spatial-level description of input data for our model of food insecurity risks in Uttar Pradesh.} \label{tab:description_of_data}
\vspace{-3mm}
  \begin{tabular}{lll}  \hline
 \textbf{District}  & \textbf{State} & \textbf{Regional} \\
\hline 
Purchasing power \cite{mb_research} &  Ration cards \cite{food_bulletin_2019, food_bulletin_2020} & Ration cards  \cite{aayog2016evaluation}  \\
Population \cite{cattaneo2021global} & Annual wheat yield & Income estimates \cite{aayog2016evaluation} \\
  & Wheat storage  \cite{food_bulletin_2019, food_bulletin_2020} &  \\
    & Average family size  \cite{food_bulletin_2019, food_bulletin_2020} &   \\
     & Drive time distances   &  \\
  \hline
  \end{tabular}}
\end{table}

\subsection{Ration Card Estimates at the District Level}\label{sect:sd_population}

The government of India issues ration cards to its citizens who are eligible to purchase subsidized grain.    
There are two types of ration cards under the NFSA: Antyodaya Anna Yojana (AAY) cards, which are intended for India's poorest citizens, and Priority Household cards, which are more common \cite{balani2013functioning}.  
Each ration card type allows different amounts of grain to be purchased at the subsidized rates.  
Specifically, AAY households can purchase 35 kg per month, while members of Priority households may buy 5 kg per month per person.  
To our knowledge, the AAY and Priority cardholder populations are not publicly available at the district level, so we estimate them.  

\begin{figure}[H]
\hspace*{-1.7cm}
\includegraphics[width=7.8cm]{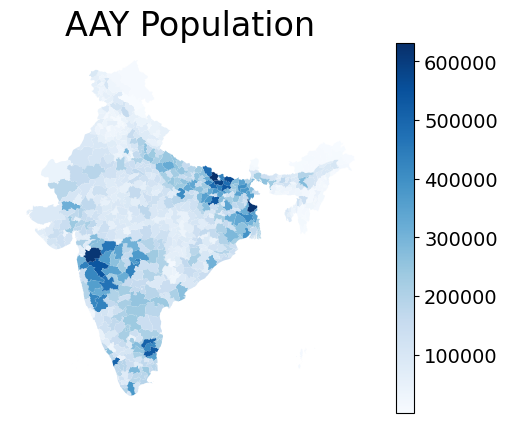}\includegraphics[width=7.5cm]{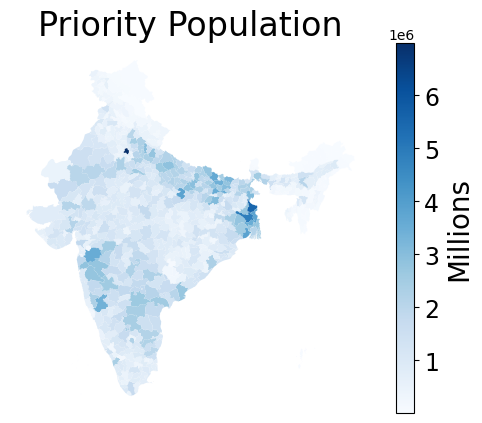}
\caption{According to India's National Food Security Act \cite{NFSA_2013}, the poorest (AAY households) receive 35 kg of grain each month, and the less poor (Priority households) receive 5 kg per person per month.}\label{fig:ration_card_estimates}
\end{figure}

To make these population estimates, we assume that the ratio of the proportion of AAY and Priority households within each district of Uttar Pradesh at the time of the 2010-2011 Census is roughly the same as today.  
Then Figure~\ref{fig:ration_card_estimates} is obtained by the following steps: 
\begin{enumerate} 
\item 
Estimate the fraction of rural (respectively urban) AAY households at a district level based on per-capita income.\footnote{See Table A6 and A7 of \cite{aayog2016evaluation}.}  
\item Use district level rural (respectively urban) population estimates to approximate the population of such AAY cardholders at a district level.\footnote{See \cite{cattaneo2021global}.}  
\item Repeat the above steps for Priority cardholders.\footnote{Note: at the time of the 2010-2011 Census, the landmark National Food Security Act (2013) had not yet been passed.  
So Tables A6 and A7 of \cite{aayog2016evaluation} provide Below Poverty Line and Above Poverty Line columns from the system at the type, the Targeted Public Distribution System.  
We use Below Poverty Line numbers for our Priority estimates.}  
\item Fill in missing rural (respectively urban) Priority and AAY cardholder population estimates at the district level by averaging the respective values from the district's neighbors.   
\item Sum the estimated Priority and AAY cardholders for the rural (respectively urban) regions.  
This gives an estimate for the total cardholders in the rural and urban regions of each district.  
\item 
Scale the district level populations so that their aggregate matches the state level estimates for rural regions provided by the Food Grain Bulletin \cite{food_bulletin_2019}.
This yields an improved estimate for the rural and urban cardholders at the district level.  
\item Now sum the rural and urban AAY (resp. Priority) cardholder population estimates, which produces an estimate of the AAY and Priority populations at the district level. 
\item Finally, due to the imprecise nature of the estimates, in a small number of districts, the ration card population estimates from the previous step exceeded the actual population of those districts, which is impossible.  
We distribute the excess ration cards from these districts uniformly among the remaining districts of the state.  
This yields the district level estimate of the AAY and Priority populations in Figure~\ref{fig:ration_card_estimates}, which agrees with ground truth when aggregated up to the state level.  
\end{enumerate}  

\subsection{Model of the PDS}\label{sect:sd_model_pds}
\begin{figure}
\hspace*{3.2cm} 
\begin{tikzpicture}
\node[circle,thick, draw = black, text = black, minimum size=0.3cm] (0) at  (2.2,2) {};
\node[circle,thick, draw = black, text = black, minimum size=0.3cm] (1) at  (3.2, 1.3)  {};
\node[circle,thick, draw = black, text = black, minimum size=0.3cm] (2) at  (4.5, 2.1)    {};
\node[circle,thick, draw = black, text = black, minimum size=0.3cm] (3) at  (2, -0.2)   {};
\node[circle,thick, draw = black, text = black, minimum size=0.3cm] (4) at  (4.3, -0.3)   {};

\draw [very thick, -stealth, latex'-latex'] (1) -- (0);
\draw [very thick, -stealth, latex'-latex'] (1) -- (4);
\draw [very thick, -stealth, latex'-latex'] (1) -- (2);
\draw [very thick, -stealth, latex'-latex'] (1) -- (3);

\node[below=0.8cm of 4] (4A) {};
\node[right=0.8cm of 4] (4B) {};
\node[left=0.8cm of 4] (4C) {};
\node[above=0.8cm of 4] (4D) {};
\draw [very thick, dash dot, -stealth,   latex'-latex'] (4) -- (4A);
\draw [very thick, dash dot, -stealth, latex'-latex'](4) -- (4B);
\draw [very thick, dash dot, -stealth, latex'-latex'] (4) -- (4C);
\draw [very thick, dash dot, -stealth, latex'-latex'] (4) -- (4D);

\node[below=0.8cm of 3] (3A) {};
\node[right=0.8cm of 3] (3B) {};
\node[left=0.8cm of 3] (3C) {};
\node[above=0.8cm of 3] (3D) {};
\draw [very thick, dash dot, -stealth, latex'-latex'] (3) -- (3A);
\draw [very thick, dash dot, -stealth, latex'-latex'] (3) -- (3B);
\draw [very thick, dash dot, -stealth, latex'-latex'] (3) -- (3C);
\draw [very thick, dash dot, -stealth, latex'-latex'] (3) -- (3D);

\node[below=0.8cm of 2] (2A) {};
\node[right=0.8cm of 2] (2B) {};
\node[left=0.8cm of 2] (2C) {};
\node[above=0.8cm of 2] (2D) {};
\draw [very thick, dash dot, -stealth, latex'-latex'] (2) -- (2A);
\draw [very thick, dash dot, -stealth, latex'-latex'] (2) -- (2B);
\draw [very thick, dash dot, -stealth, latex'-latex'] (2) -- (2C);
\draw [very thick, dash dot, -stealth, latex'-latex'] (2) -- (2D);

\node[below=0.75cm of 0] (0A) {};
\node[right=0.75cm of 0] (0B) {};
\node[left=0.75cm of 0] (0C) {};
\node[above=0.75cm of 0] (0D) {};
\draw [very thick, dash dot, -stealth, latex'-latex'] (0) -- (0A);
\draw [very thick, dash dot, -stealth, latex'-latex'] (0) -- (0B);
\draw [very thick, dash dot, -stealth, latex'-latex'] (0) -- (0C);
\draw [very thick, dash dot, -stealth, latex'-latex'] (0) -- (0D);

\end{tikzpicture}
\caption{
Our model for the PDS of Uttar Pradesh is given by a fully connected, bidirectional directed temporal graph, 
where nodes are districts and edges are transportation routes between those districts.  
The bi-directional dotted edges are a short-hand way to depict that every node is connected to every other node.  
The same differential equations given by a stock and flow diagram determine the trajectory of wheat within all nodes,  
and import requests and surplus storage among the districts determine the flow of wheat between edges.  
}\label{fig:PDS_model}
\end{figure}
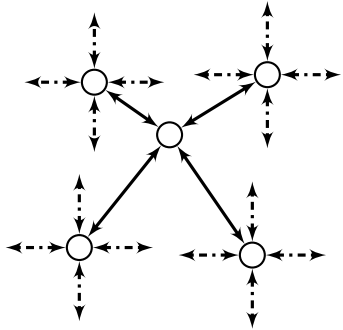

We model the PDS of grain in Uttar Pradesh as a fully connected, bidirectional graph with districts as nodes and transportation routes as edges.  See Figure~\ref{fig:PDS_model}.  

\subsubsection{Nodes} 

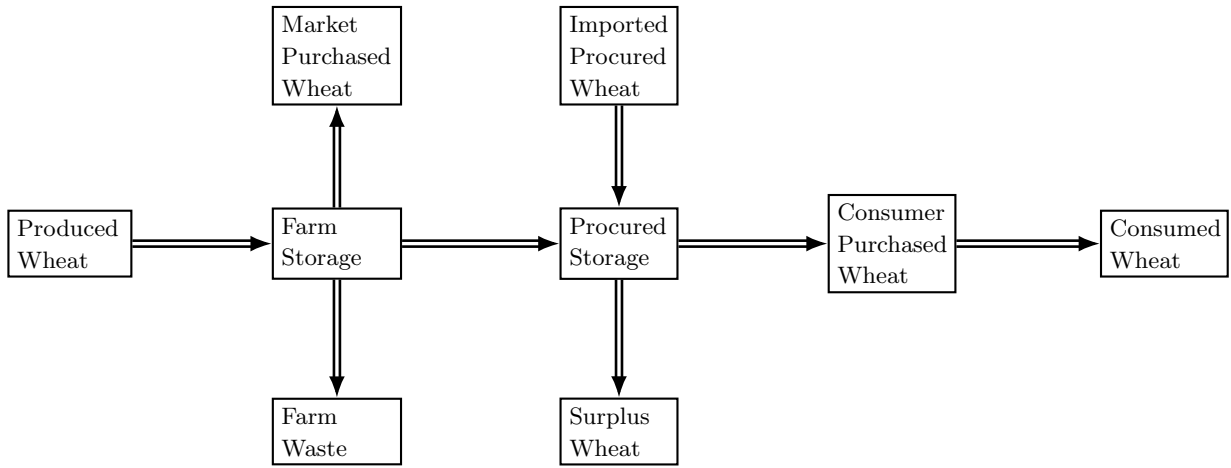
\begin{figure}[H]
\hspace*{-2.12cm} 
\begin{tikzpicture}
\node[rectangle,thick, draw = black, text = black, minimum size=0.3cm, text width=1.4cm] (0) at  (0.2, 0) {\small Produced Wheat};
\node[rectangle,thick, draw = black, text = black, minimum size=0.3cm, text width=1.47cm] (1) at  (3.75, 0)  {\small Farm Storage};
\node[rectangle,thick, draw = black, text = black, minimum size=0.3cm, text width=1.47cm] (2) at  (3.75, -2.5)  {\small Farm Waste};
\node[rectangle,thick, draw = black, text = black, minimum size=0.3cm, text width=1.47cm] (3) at  (3.75, 2.5)  {\small Market Purchased Wheat}; 
\node[rectangle,thick, draw = black, text = black, minimum size=0.3cm, text width=1.32cm] (4) at  (7.5, 2.5)    {\small Imported Procured Wheat};
\node[rectangle,thick, draw = black, text = black, minimum size=0.3cm, text width=1.32cm] (5) at  (7.5, 0)    {\small Procured Storage};
\node[rectangle,thick, draw = black, text = black, minimum size=0.3cm, text width=1.32cm] (6) at  (7.5, -2.5)    {\small Surplus Wheat};
\node[rectangle,thick, draw = black, text = black, minimum size=0.3cm, text width=1.45cm] (7) at  (11.13, 0)   {\small Consumer \\ Purchased Wheat};
\node[rectangle,thick, draw = black, text = black, minimum size=0.3cm, text width=1.45cm] (8) at  (14.75, 0)   {\small Consumed Wheat};

\draw[line width=1pt, double distance=1.25pt, arrows = {-Latex[length=0pt 3 0.25]}] (0) -- (1); 
\draw[line width=1pt, double distance=1.25pt, arrows = {-Latex[length=0pt 3 0.25]}] (1) -- (2);
\draw[line width=1pt, double distance=1.25pt, arrows = {-Latex[length=0pt 3 0.25]}] (1) -- (3);
\draw[line width=1pt, double distance=1.25pt, arrows = {-Latex[length=0pt 3 0.25]}] (1) -- (5);
\draw[line width=1pt, double distance=1.25pt, arrows = {-Latex[length=0pt 3 0.25]}] (4) -- (5);
\draw[line width=1pt, double distance=1.25pt, arrows = {-Latex[length=0pt 3 0.25]}] (5) -- (6);
\draw[line width=1pt, double distance=1.25pt, arrows = {-Latex[length=0pt 3 0.25]}] (5) -- (7);
\draw[line width=1pt, double distance=1.25pt, arrows = {-Latex[length=0pt 3 0.25]}] (7) -- (8); 
\end{tikzpicture} 
\caption{When districts produce wheat, wheat flows from the Produced Wheat stock to the Farm Storage stock.  
The market purchases some portion of that wheat, which it collects in the Market Purchased Grain stock.  
Excess wheat flows to Farm Waste and is no longer used in the model.  
The remaining wheat flows to the Procured Storage stock, which holds wheat purchased by the state or national government that the district can use over the coming weeks.  
Any excess wheat flows to the Surplus Wheat stock, where it is available for transportation to other districts within the state.  
Ration card population estimates together with state-wide wheat consumption estimates determine each district's weekly wheat requirements.  
Then available wheat in the Procured Storage stock flows to the Consumer Purchased Wheat stock and ultimately arrives in the Consumed Wheat stock.  
Districts that do not have enough wheat in the Procured Storage stock for the coming weeks request it from districts that have a surplus.  
Any requested wheat arrives in the Imported Procured Wheat stock, where it subsequently flows to the Procured Storage stock to be available to that district's consumers.  
We provide the complete system dynamics diagram in the appendix, which has stocks, flows and information arrows.}\label{fig:vensim_model}
\end{figure}

\begin{figure}[H]
\hspace{-1.7cm}
\includegraphics[width=15cm]{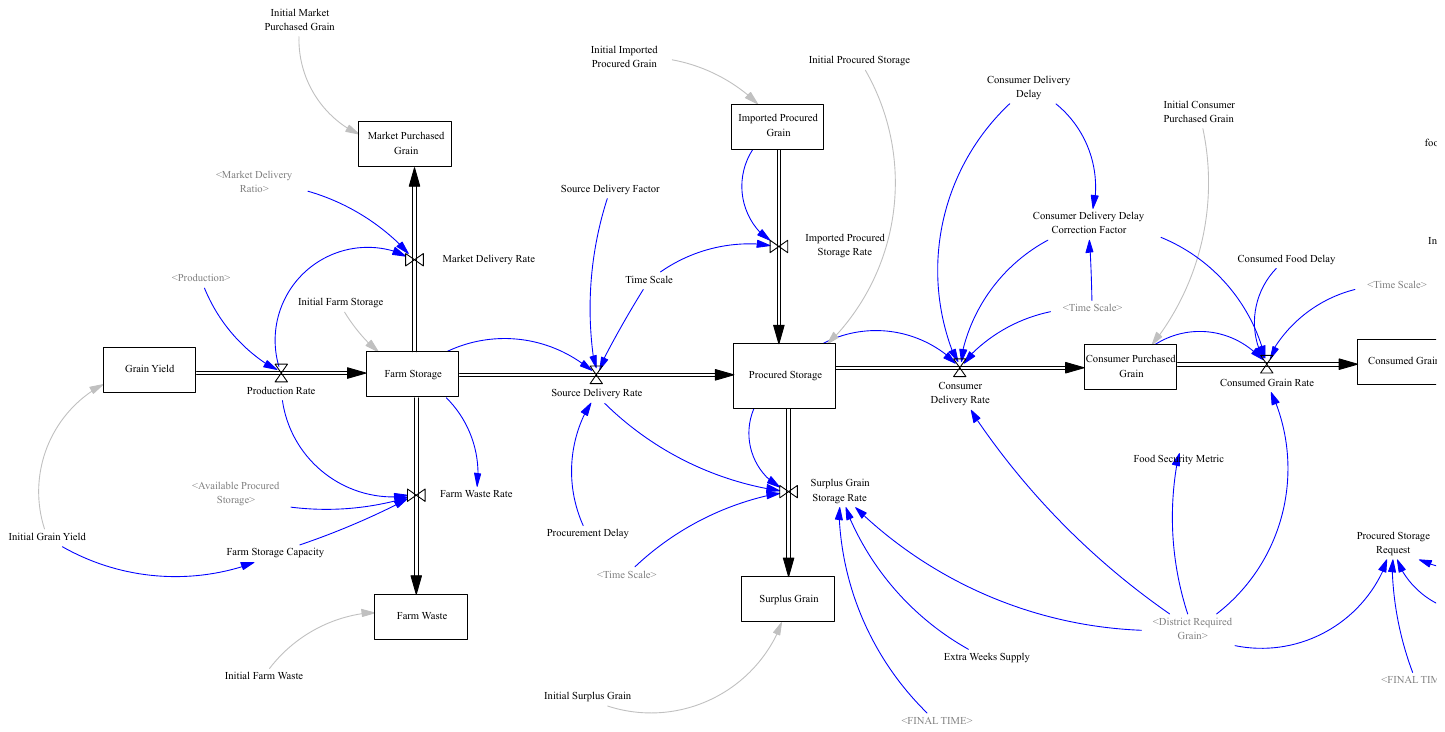}
\caption{This is an image of the main components of our stock and flow model implemented in Vensim.}\label{fig:vensim}
\end{figure}

All nodes of Figure~\ref{fig:PDS_model} are represented by the same model of the PDS at the district level, which Figure~\ref{fig:vensim_model} describes.  This model is implemented in Vensim, see Figure~\ref{fig:vensim}.  

The stock and flow diagram of Figure~\ref{fig:vensim_model} begins at the Produced Wheat stock.  
Starting from an initial district level distribution of produced wheat in Uttar Pradesh, 
we scale those values so that the aggregate sum of all produced wheat will match our ground truth value of 32.6 million metric tonnes of wheat from 2018.  

From the Farm Storage stock, wheat flows to either Procured Storage, Farm Waste or the Market Purchased Wheat stock.  
We use the following expression to describe how much wheat enters the Market Purchased Wheat stock. 
\begin{align} 
&(\textrm{Last Year's Non-Wasted Wheat Harvest} - \textrm{Last Year's Procured Wheat}) \label{eq:last_years_market} \\ 
& \quad  \quad  \times \frac{\textrm{MSP}}{\textrm{Last Year's MSP}} \times \frac{\textrm{Last Year's Market Price}}{\textrm{Market Price}}\,. \label{eq:change_prices}
\end{align} 
Observe that \eqref{eq:last_years_market} is the amount of wheat that the market purchased last year.  
Then this year's estimate of market purchased wheat is positively correlated with Equation~\eqref{eq:change_prices}.  
Notice that if MSP prices increase more than the market price year-over-year, i.e., 
\[
\eqref{eq:change_prices} > 1\,,
\] 
then the model predicts less wheat will enter the Market Procured Wheat stock this year, 
and therefore more will enter the Procured Storage stock.  
This agrees with intuition, because farmers would be incentivized to sell more wheat to state and national governments.  
The reverse occurs if the market price of wheat increases more than the MSP.  

From the Procured Storage stock, wheat flows to the Consumed Wheat stock or the Surplus Storage stock depending on whether the district has enough wheat for the coming weeks.  
The rate at which each district consumes wheat does not seem to be publicly available, so we estimate it using our district level ration card population estimates obtained in Section~\ref{sect:sd_population}. 
We then scale these district level estimates so that their aggregate matches the rate that wheat is depleted from the Procured Storage levels at the state-level, for which we have ground truth \cite{food_bulletin_2019, food_bulletin_2020}.  
This determines the rate that each district consumes wheat in one week.  

The model transports wheat from a district with positive Surplus Storage to a district that requests it.  
Each district tries to maintain a four week supply of wheat in their Procured Storage, so those without this reserve will request more wheat.  
Districts receive grain through the Imported Procured Wheat stock.  

\subsubsection{Edges} 
Wheat transportation data between districts in Uttar Pradesh also do not seem to be publicly available, so we assume that any district can transfer wheat to any other district.  See Figure~\ref{fig:PDS_model}.  
We also assume that districts transport wheat to closer districts before farther districts, where distance was measured in terms of a drive time estimate between the largest population center of each district.  

\subsection{Results}\label{sect:sd_results}

To model food insecurity risk, we first estimate typical undernourishment rates among the districts of Uttar Pradesh.  
Our estimates come from two sources.  
The first source is Table~2 of \cite{menon2008comparisons}, which is a 2009 report that estimates the percent undernourished in 17 states in India. The percent undernourished wasas defined by the 2008 Global Hunger Index as consuming less than 1,632 kcal per day.  

\begin{figure}
\hspace*{-1cm}
\begin{subfigure}[b]{0.3\textwidth} 
\includegraphics[width=6.5cm]{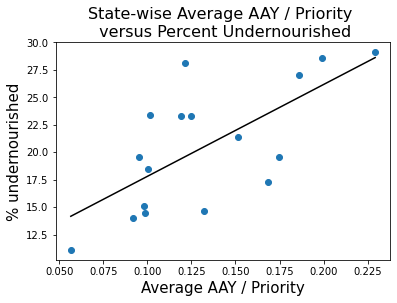}
\caption{}\label{fig:scatter_undernourished}
\end{subfigure}
\hspace*{5.4cm}
\begin{subfigure}[b]{0.3\textwidth}
\hspace*{-1.9cm}
\includegraphics[width=6.5cm]{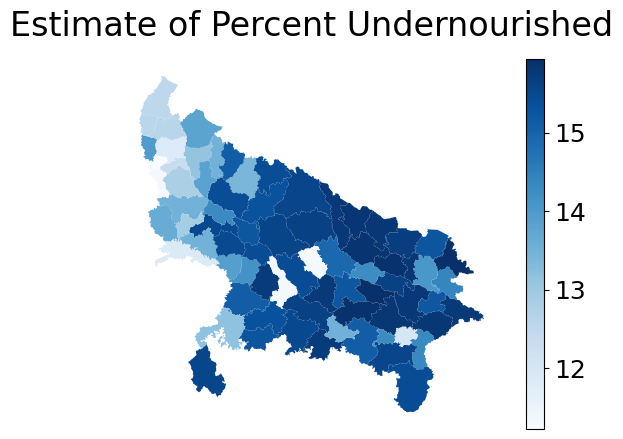}
\caption{}\label{fig:map_undernourished}
\end{subfigure}
\caption{
There is ground truth of the percent of citizens who are undernourished at the state level, but not at the district level.  
We need this information at the district level for our model, so we estimate it.  
Figure~\ref{fig:scatter_undernourished} is a scatter plot of the relationship between a state's percent undernourished and its ratio of AAY to Priority populations.  
The correlation of these two variables is strongly statistically significant.  
The line in Figure~\ref{fig:scatter_undernourished} is the best fit line by least squares linear regression, which has a slope of 83.67 and a p-value of $3.67\mathrm{e}{-149}$.  
Figure~\ref{fig:map_undernourished} is the estimate of each district's percent undernourished that results from computing the ratio of the estimated AAY and Priority populations at the district level (see Figure~\ref{fig:ration_card_estimates}) and then using the linear regression line from Figure~\ref{fig:scatter_undernourished}.  
}\label{fig:undernourished_estimate}
\end{figure}

The second source is the ratio of the AAY population to the Priority population among these states.  
Figure~\ref{fig:scatter_undernourished} shows these quantities are correlated, 
which is intuitive, because a greater proportion of AAY residents in a population should mean that a greater proportion of those residents need food assistance.  
We assume that this relationship, given by the slope of the line in Figure~\ref{fig:scatter_undernourished}, also holds for the districts of Uttar Pradesh.  
An estimate of the percent undernourished in Uttar Pradesh can then be derived using our district level estimates of the AAY and Priority populations.  
See Figure~\ref{fig:map_undernourished}.  

\begin{figure}
\hspace*{-1cm}
\begin{subfigure}[b]{0.3\textwidth}
\includegraphics[width=6.5cm]{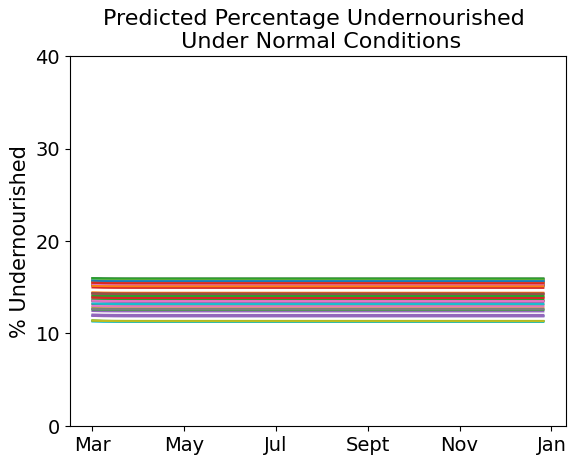} 
\caption{}\label{fig:pred_undernourished_normal}
\end{subfigure}
\hspace*{5.4cm}
\begin{subfigure}[b]{0.3\textwidth}
\hspace*{-1.9cm}
\includegraphics[width=6.5cm]{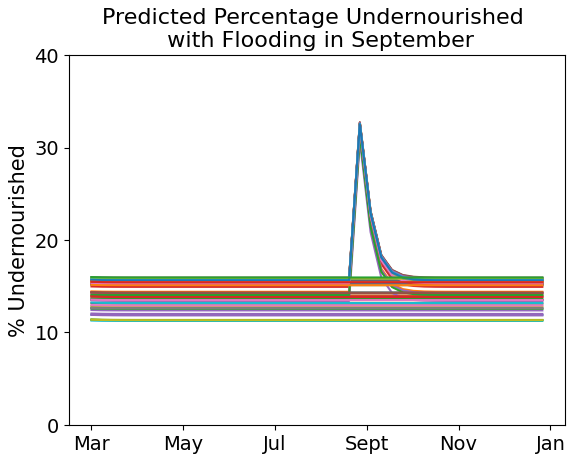}
\caption{}\label{fig:prediction_flooding}
\end{subfigure}
\caption{The 75 curves represent the predicted percent undernourished for the 75 districts in Uttar Pradesh in 2019.  
Figure~\ref{fig:pred_undernourished_normal} is the predicted undernourishment for each district under normal conditions.  
Figure~\ref{fig:prediction_flooding} includes a September flooding event in northeast Uttar Pradesh, where 75\% of the stored wheat in 21 districts is destroyed.  
The model predicts a temporary spike in food insecurity for those districts which is soon restored. }\label{fig:prediction_normal}
\end{figure}

Because our estimates are based on incomplete data, our model does not predict the exact percent of undernourished, but rather it predicts trends in undernourishment.  
Figure~\ref{fig:pred_undernourished_normal} is the model's prediction under normal circumstances.  
Figure~\ref{fig:prediction_flooding} is the model's prediction under a flooding event in September. In this scenario, there is a temporary spike in undernourishment in the affected districts. The model predicts that wheat from neighboring districts will re-establish normal food insecurity levels for those impacted by the flood.  

\begin{figure}[H]
\hspace*{-1cm}
\begin{subfigure}[b]{0.3\textwidth}
\includegraphics[width=6.5cm]{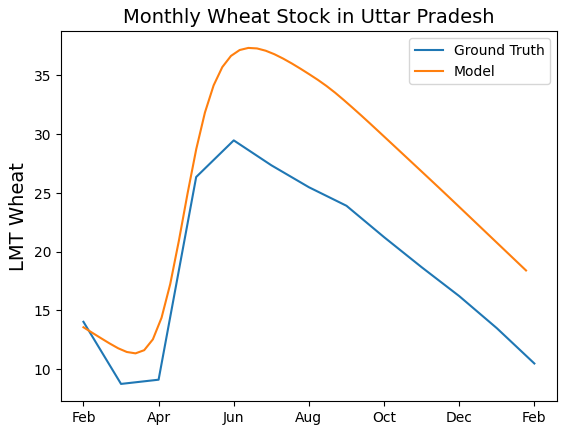}
\caption{}\label{fig:2019_calibration}
\end{subfigure}
\hspace*{5.4cm}
\begin{subfigure}[b]{0.3\textwidth}
\hspace*{-1.9cm}
\includegraphics[width=6.5cm]{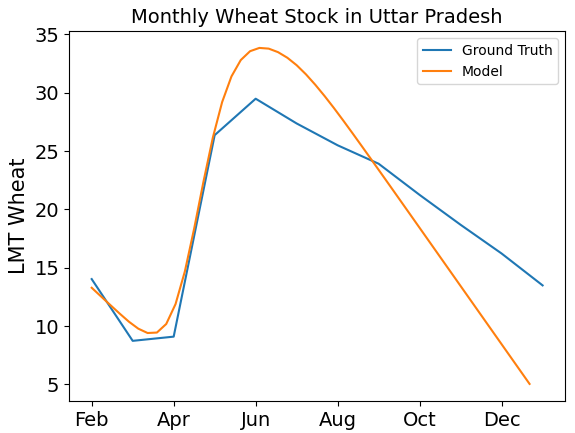}
\caption{}\label{fig:2018_data_predict_2019}
\end{subfigure}
\caption{These plots show the ground truth wheat storage per month in Uttar Pradesh in 2019 as well as the model's prediction after aggregating from the district level up to the state level. 
Figure~\ref{fig:2019_calibration} uses 2019 data to predict the rate that wheat leaves the system in 2019.  
Figure~\ref{fig:2018_data_predict_2019} uses 2018 data for 2019 predictions.  
}\label{fig:estimated_wheat_state}
\end{figure}

The only ground truth input to the model that is not temporally sparse is the amount of wheat in storage per month at the state level. 
Figure~\ref{fig:estimated_wheat_state} compares the ground truth wheat in storage at the state level against the aggregate over all districts of the model's predicted wheat in storage.  
The two curves should trend together.  
Figure~\ref{fig:2019_calibration} calibrates the model by using 2019 inputs to predict wheat storage in 2019.  
Overall, the model fits the ground truth well, but it predicts too much grain enters the PDS after the spring harvest.  
This may be improved with a more accurate model of the market dynamics that determine whether wheat enters into the PDS or the public market (see Equation~\eqref{eq:change_prices}).    
Figure~\ref{fig:2018_data_predict_2019} describes a realistic scenario where only data from previous years is available.  
The model again predicts that too much wheat enters the PDS, but this time it also predicts that wheat leaves the system too quickly.  
Regarding the latter property, the model uses ground truth data on the rate the wheat left the system in 2018 as the rate in 2019, which might be improved by instead averaging rates over several previous years.  

\section{Conclusion} 
The global challenge of food insecurity, exacerbated by conflicts such as the war in Ukraine and by climate change-induced events like heatwaves and droughts, underscores the urgent need for innovative solutions to ensure food security for all. Focusing our research on the wheat breadbasket of northern India, a critical region responsible for sustaining not only India's vast population but also serving as a significant source of grain to the rest of the world, we have outlined a comprehensive framework to address the complexities of crop production, distribution, and resilience in the face of evolving environmental and socio-economic factors.

Through the integration of satellite remote sensing, deep learning classification, physics-based crop yield simulations, and system dynamics modeling, our research offers a multifaceted approach to understanding and improving the dynamics of food production and distribution systems. By developing predictive models for crop production under diverse climate scenarios and simulating the distribution of grain to food-insecure populations, we aim to provide actionable insights that can inform policy decisions and interventions aimed at enhancing the resilience of food systems.

While our focus has been on the wheat production landscape of northern India, we acknowledge the challenges posed by the lack of ground truth data for validation purposes. Thus, we have leveraged curated datasets from regions like France to validate our algorithms, while tailoring our distribution and market dynamics models to the unique characteristics of the Indian economy and governance structure.

In conclusion, this research represents a significant step towards harnessing technology and data-driven approaches to address the complex and interconnected challenges of food security. By fostering collaboration between stakeholders, policymakers, and researchers, we can work towards building more resilient and sustainable food systems that ensure equitable access to nutritious food for all, even in the face of global disruptions.
\section{Acknowledgments}
NASA POWER data were obtained from the National Aeronautics and Space Administration (NASA) Langley Research Center (LaRC) Prediction of Worldwide Energy Resource (POWER) Project funded through the NASA Earth Science/Applied Science Program.

The authors would like to thank the following members of our team who contributed to this research: Monica Barbu-McInnis, Anuraag Kaashyap, Heather Phelps, Dan Mauer, Anneliese Braunegg, Meryl Flaherty, and Mark Zimmermann. This work was funded by MITRE's internal research and development program.

\section{Appendix}
This is the detailed stock and flow diagram for each node of our model for the PDS of Uttar Pradesh.  \\ 

\bibliography{food_insecurity}

\begin{thebibliography}{39}
\providecommand{\natexlab}[1]{#1}
\providecommand{\url}[1]{#1}
\csname url@samestyle\endcsname
\providecommand{\newblock}{\relax}
\providecommand{\bibinfo}[2]{#2}
\providecommand{\BIBentrySTDinterwordspacing}{\spaceskip=0pt\relax}
\providecommand{\BIBentryALTinterwordstretchfactor}{4}
\providecommand{\BIBentryALTinterwordspacing}{\spaceskip=\fontdimen2\font plus
\BIBentryALTinterwordstretchfactor\fontdimen3\font minus
  \fontdimen4\font\relax}
\providecommand{\BIBforeignlanguage}[2]{{%
\expandafter\ifx\csname l@#1\endcsname\relax
\typeout{** WARNING: IEEEtranN.bst: No hyphenation pattern has been}%
\typeout{** loaded for the language `#1'. Using the pattern for}%
\typeout{** the default language instead.}%
\else
\language=\csname l@#1\endcsname
\fi
#2}}
\providecommand{\BIBdecl}{\relax}
\BIBdecl

\bibitem[FAO(2023)]{FAO2023}
\BIBentryALTinterwordspacing
2023. [Online]. Available:
  \url{https://www.fao.org/3/cc3017en/online/cc3017en.html}
\BIBentrySTDinterwordspacing

\bibitem[Kumar(2022)]{Kumar_2022}
\BIBentryALTinterwordspacing
T.~N. Kumar, ``Lessons for today from india’s 2006 wheat crisis,'' May 2022.
  [Online]. Available:
  \url{https://indianexpress.com/article/opinion/columns/lessons-for-today-from-indias-2006-wheat-crisis-grain-export-ban-7924531/}
\BIBentrySTDinterwordspacing

\bibitem[ind()]{indiacensus}
\BIBentryALTinterwordspacing
 [Online]. Available: \url{https://agcensus.gov.in/AgriCensus/}
\BIBentrySTDinterwordspacing

\bibitem[Nakalembe and Kerner(2023)]{nakalembe2023considerations}
C.~Nakalembe and H.~Kerner, \emph{Considerations for AI-EO for agriculture in
  Sub-Saharan Africa}.\hskip 1em plus 0.5em minus 0.4em\relax Institute of
  Physics, 2023.

\bibitem[Ru{\ss}wurm et~al.(2019)Ru{\ss}wurm, Lef{\`{e}}vre, and
  K{\"{o}}rner]{breizhcrops}
\BIBentryALTinterwordspacing
M.~Ru{\ss}wurm, S.~Lef{\`{e}}vre, and M.~K{\"{o}}rner, ``Breizhcrops: {A}
  satellite time series dataset for crop type identification,'' \emph{CoRR},
  vol. abs/1905.11893, 2019. [Online]. Available:
  \url{http://arxiv.org/abs/1905.11893}
\BIBentrySTDinterwordspacing

\bibitem[of~Geographic and Information(2020)]{rpg_dataset}
\BIBentryALTinterwordspacing
F.~N.~I. of~Geographic and F.~Information. (2020) Rpg crop type parcel data.
  [Online]. Available:
  \url{https://www.data.gouv.fr/en/datasets/registre-parcellaire-graphique-rpg-contours-des-parcelles-et-ilots-culturaux-et-leur-groupe-de-cultures-majoritaire/}
\BIBentrySTDinterwordspacing

\bibitem[Skakun et~al.(2022)Skakun, Wevers, Brockmann, Doxani, Aleksandrov,
  Bati{\v{c}}, Frantz, Gascon, G{\'o}mez-Chova, Hagolle,
  et~al.]{skakun2022cloud}
S.~Skakun, J.~Wevers, C.~Brockmann, G.~Doxani, M.~Aleksandrov, M.~Bati{\v{c}},
  D.~Frantz, F.~Gascon, L.~G{\'o}mez-Chova, O.~Hagolle \emph{et~al.}, ``Cloud
  mask intercomparison exercise (cmix): An evaluation of cloud masking
  algorithms for landsat 8 and sentinel-2,'' \emph{Remote Sensing of
  Environment}, vol. 274, p. 112990, 2022.

\bibitem[Ronneberger et~al.(2015)Ronneberger, Fischer, and Brox]{unet}
\BIBentryALTinterwordspacing
O.~Ronneberger, P.~Fischer, and T.~Brox, ``U-net: Convolutional networks for
  biomedical image segmentation,'' \emph{CoRR}, vol. abs/1505.04597, 2015.
  [Online]. Available: \url{http://arxiv.org/abs/1505.04597}
\BIBentrySTDinterwordspacing

\bibitem[He et~al.(2015)He, Zhang, Ren, and Sun]{resnet}
\BIBentryALTinterwordspacing
K.~He, X.~Zhang, S.~Ren, and J.~Sun, ``Deep residual learning for image
  recognition,'' 2015. [Online]. Available:
  \url{https://arxiv.org/abs/1512.03385}
\BIBentrySTDinterwordspacing

\bibitem[Yuan et~al.(2021)Yuan, Shi, and Gu]{rs_model_survey}
\BIBentryALTinterwordspacing
X.~Yuan, J.~Shi, and L.~Gu, ``A review of deep learning methods for semantic
  segmentation of remote sensing imagery,'' \emph{Expert Systems with
  Applications}, vol. 169, p. 114417, 2021. [Online]. Available:
  \url{https://www.sciencedirect.com/science/article/pii/S0957417420310836}
\BIBentrySTDinterwordspacing

\bibitem[M~Rustowicz et~al.(2019)M~Rustowicz, Cheong, Wang, Ermon, Burke, and
  Lobell]{crop_type_africa}
R.~M~Rustowicz, R.~Cheong, L.~Wang, S.~Ermon, M.~Burke, and D.~Lobell,
  ``Semantic segmentation of crop type in africa: A novel dataset and analysis
  of deep learning methods,'' in \emph{Proceedings of the IEEE/CVF Conference
  on Computer Vision and Pattern Recognition (CVPR) Workshops}, June 2019.

\bibitem[Institute(2010)]{mapspam}
\BIBentryALTinterwordspacing
I.~F. P.~R. Institute. (2010) Mapspam. [Online]. Available:
  \url{https://www.mapspam.info/}
\BIBentrySTDinterwordspacing

\bibitem[McInnes et~al.(2018)McInnes, Healy, and Melville]{umap}
\BIBentryALTinterwordspacing
L.~McInnes, J.~Healy, and J.~Melville, ``Umap: Uniform manifold approximation
  and projection for dimension reduction,'' 2018. [Online]. Available:
  \url{https://arxiv.org/abs/1802.03426}
\BIBentrySTDinterwordspacing

\bibitem[Sayre et~al.(2014)Sayre, Dangermond, Frye, Vaughan, Aniello, Breyer,
  Cribbs, Hopkins, Naumann, Derrenbacher, Wright, Brown, Butler, Bennett,
  Smith, Benson, Sistine, Warner, Cress, and Grosse]{usgs_elu}
R.~Sayre, J.~Dangermond, C.~Frye, R.~Vaughan, P.~Aniello, S.~Breyer, D.~Cribbs,
  D.~Hopkins, R.~Naumann, B.~Derrenbacher, D.~Wright, C.~Brown, K.~Butler,
  L.~Bennett, J.~Smith, L.~Benson, D.~Sistine, H.~Warner, J.~Cress, and
  A.~Grosse, \emph{A New Map of Global Ecological Land Units — An
  Ecophysiographic Stratification Approach.}, 12 2014.

\bibitem[Chen et~al.(2020)Chen, Kornblith, Norouzi, and
  Hinton]{contrastive_learning}
\BIBentryALTinterwordspacing
T.~Chen, S.~Kornblith, M.~Norouzi, and G.~Hinton, ``A simple framework for
  contrastive learning of visual representations,'' 2020. [Online]. Available:
  \url{https://arxiv.org/abs/2002.05709}
\BIBentrySTDinterwordspacing

\bibitem[Zhang et~al.(2021)Zhang, Zhang, Zhang, Chen, Wang, and Wen]{proda}
\BIBentryALTinterwordspacing
P.~Zhang, B.~Zhang, T.~Zhang, D.~Chen, Y.~Wang, and F.~Wen, ``Prototypical
  pseudo label denoising and target structure learning for domain adaptive
  semantic segmentation,'' in \emph{{IEEE} Conference on Computer Vision and
  Pattern Recognition, {CVPR} 2021, virtual, June 19-25, 2021}.\hskip 1em plus
  0.5em minus 0.4em\relax Computer Vision Foundation / {IEEE}, 2021, pp.
  12\,414--12\,424. [Online]. Available:
  \url{https://openaccess.thecvf.com/content/CVPR2021/html/Zhang\_Prototypical\_Pseudo\_Label\_Denoising\_and\_Target\_Structure\_Learning\_for\_Domain\_CVPR\_2021\_paper.html}
\BIBentrySTDinterwordspacing

\bibitem[Vu et~al.(2019)Vu, Jain, Bucher, Cord, and P{\'{e}}rez]{advent}
\BIBentryALTinterwordspacing
T.~Vu, H.~Jain, M.~Bucher, M.~Cord, and P.~P{\'{e}}rez, ``{ADVENT:} adversarial
  entropy minimization for domain adaptation in semantic segmentation,'' in
  \emph{{IEEE} Conference on Computer Vision and Pattern Recognition, {CVPR}
  2019, Long Beach, CA, USA, June 16-20, 2019}.\hskip 1em plus 0.5em minus
  0.4em\relax Computer Vision Foundation / {IEEE}, 2019, pp. 2517--2526.
  [Online]. Available:
  \url{http://openaccess.thecvf.com/content\_CVPR\_2019/html/Vu\_ADVENT\_Adversarial\_Entropy\_Minimization\_for\_Domain\_Adaptation\_in\_Semantic\_Segmentation\_CVPR\_2019\_paper.html}
\BIBentrySTDinterwordspacing

\bibitem[de~Wit(2024)]{de2024pcse}
A.~de~Wit, ``Pcse documentation,'' Tech. Rep., 2019.[Online]. Available, Tech.
  Rep., 2024.

\bibitem[Van~Diepen et~al.(1989)Van~Diepen, Wolf, Van~Keulen, and
  Rappoldt]{van1989wofost}
C.~v. Van~Diepen, J.~v. Wolf, H.~Van~Keulen, and C.~Rappoldt, ``Wofost: a
  simulation model of crop production,'' \emph{Soil use and management},
  vol.~5, no.~1, pp. 16--24, 1989.

\bibitem[Nachtergaele et~al.(2009)Nachtergaele, Velthuizen, Verelst, and
  Wiberg]{hwsd1}
F.~Nachtergaele, H.~Velthuizen, L.~Verelst, and D.~Wiberg, ``Harmonized world
  soil database (hwsd),'' \emph{Food and Agriculture Organization of the United
  Nations, Rome}, 2009.

\bibitem[Nachtergaele et~al.(2023)Nachtergaele, van Velthuizen, Verelst,
  Wiberg, Henry, Chiozza, Yigini, Aksoy, Batjes, Boateng, et~al.]{hwsd2}
F.~Nachtergaele, H.~van Velthuizen, L.~Verelst, D.~Wiberg, M.~Henry,
  F.~Chiozza, Y.~Yigini, E.~Aksoy, N.~Batjes, E.~Boateng \emph{et~al.},
  \emph{Harmonized World Soil Database version 2.0}.\hskip 1em plus 0.5em minus
  0.4em\relax Food and Agriculture Organization of the United Nations, 2023.

\bibitem[Jones and Thornton(2000)]{jones2000marksim}
P.~G. Jones and P.~K. Thornton, ``Marksim: software to generate daily weather
  data for latin america and africa,'' \emph{Agronomy Journal}, vol.~92, no.~3,
  pp. 445--453, 2000.

\bibitem[Ben-Ari et~al.(2018)Ben-Ari, Bo{\'e}, Ciais, Lecerf, Van~der Velde,
  and Makowski]{ben2018causes}
T.~Ben-Ari, J.~Bo{\'e}, P.~Ciais, R.~Lecerf, M.~Van~der Velde, and D.~Makowski,
  ``Causes and implications of the unforeseen 2016 extreme yield loss in the
  breadbasket of france,'' \emph{Nature communications}, vol.~9, no.~1, p.
  1627, 2018.

\bibitem[N{\'o}ia~J{\'u}nior et~al.(2023)N{\'o}ia~J{\'u}nior, Deswarte, Cohan,
  Martre, van Der~Velde, Lecerf, Webber, Ewert, Ruane, Slafer,
  et~al.]{noia2023extreme}
R.~d.~S. N{\'o}ia~J{\'u}nior, J.-C. Deswarte, J.-P. Cohan, P.~Martre, M.~van
  Der~Velde, R.~Lecerf, H.~Webber, F.~Ewert, A.~C. Ruane, G.~A. Slafer
  \emph{et~al.}, ``The extreme 2016 wheat yield failure in france,''
  \emph{Global Change Biology}, vol.~29, no.~11, pp. 3130--3146, 2023.

\bibitem[{van der Velde} et~al.(2020){van der Velde}, Lecerf, d’Andrimont,
  and Ben-Ari]{VANDERVELDE2020139}
\BIBentryALTinterwordspacing
M.~{van der Velde}, R.~Lecerf, R.~d’Andrimont, and T.~Ben-Ari, ``Chapter 8 -
  assessing the france 2016 extreme wheat production loss—evaluating our
  operational capacity to predict complex compound events,'' in \emph{Climate
  Extremes and Their Implications for Impact and Risk Assessment}, J.~Sillmann,
  S.~Sippel, and S.~Russo, Eds.\hskip 1em plus 0.5em minus 0.4em\relax
  Elsevier, 2020, pp. 139--158. [Online]. Available:
  \url{https://www.sciencedirect.com/science/article/pii/B9780128148952000094}
\BIBentrySTDinterwordspacing

\bibitem[Goldsmith et~al.(2018)Goldsmith, Scheipl, Huang, Wrobel, Gellar,
  Harezlak, McLean, Swihart, Xiao, Crainiceanu, and Reiss]{refund}
\BIBentryALTinterwordspacing
J.~Goldsmith, F.~Scheipl, L.~Huang, J.~Wrobel, J.~Gellar, J.~Harezlak, M.~W.
  McLean, B.~Swihart, L.~Xiao, C.~Crainiceanu, and P.~T. Reiss, \emph{refund:
  Regression with Functional Data}, 2018, r package version 0.1-17. [Online].
  Available: \url{https://CRAN.R-project.org/package=refund}
\BIBentrySTDinterwordspacing

\bibitem[Park et~al.(2023)Park, Li, and Li]{park2023crop}
Y.~Park, B.~Li, and Y.~Li, ``Crop yield prediction using bayesian spatially
  varying coefficient models with functional predictors,'' \emph{Journal of the
  American Statistical Association}, vol. 118, no. 541, pp. 70--83, 2023.

\bibitem[Dziak et~al.(2019)Dziak, Coffman, Reimherr, Petrovich, Li, Shiffman,
  and Shiyko]{dziak2019scalar}
J.~J. Dziak, D.~L. Coffman, M.~Reimherr, J.~Petrovich, R.~Li, S.~Shiffman, and
  M.~P. Shiyko, ``Scalar-on-function regression for predicting distal outcomes
  from intensively gathered longitudinal data: Interpretability for applied
  scientists,'' \emph{Statistics surveys}, vol.~13, p. 150, 2019.

\bibitem[Bhatia et~al.(1985)]{bhatia1985food}
B.~M. Bhatia \emph{et~al.}, \emph{Food security in South Asia.}\hskip 1em plus
  0.5em minus 0.4em\relax Oxford and IBH, 1985.

\bibitem[Puri(2022)]{puri2022india}
R.~Puri, ``India's national food security act (nfsa): Early experiences,''
  \emph{Food Governance in India}, pp. 1--18, 2022.

\bibitem[Pandey(2012)]{pandey2012uttar}
A.~K. Pandey, ``Uttar pradesh: State economy (at a glance),'' 2012.

\bibitem[mb_()]{mb_research}
\BIBentryALTinterwordspacing
``Mb-research internationale marktdaten.'' [Online]. Available:
  \url{https://www.english.mb-research.de/index.html}
\BIBentrySTDinterwordspacing

\bibitem[of~India(2019)]{food_bulletin_2019}
\BIBentryALTinterwordspacing
G.~of~India. (2019) Food grain bulletin. [Online]. Available:
  \url{https://dfpd.gov.in/food-grain-bulletin.htm}
\BIBentrySTDinterwordspacing

\bibitem[of~India(2020)]{food_bulletin_2020}
\BIBentryALTinterwordspacing
------. (2020) Food grain bulletin. [Online]. Available:
  \url{https://dfpd.gov.in/food-grain-bulletin.htm}
\BIBentrySTDinterwordspacing

\bibitem[Aayog(2016)]{aayog2016evaluation}
N.~Aayog, ``Evaluation study on role of public distribution system in shaping
  household and nutritional security india,'' \emph{Policy}, vol.~72, p.~80,
  2016.

\bibitem[Cattaneo et~al.(2021)Cattaneo, Nelson, and
  McMenomy]{cattaneo2021global}
A.~Cattaneo, A.~Nelson, and T.~McMenomy, ``Global mapping of urban--rural
  catchment areas reveals unequal access to services - check,''
  \emph{Proceedings of the National Academy of Sciences}, vol. 118, no.~2, p.
  e2011990118, 2021.

\bibitem[Balani(2013)]{balani2013functioning}
S.~Balani, ``Functioning of the public distribution system,'' 2013.

\bibitem[NFS(2013)]{NFSA_2013}
\BIBentryALTinterwordspacing
(2013). [Online]. Available: \url{https://dfpd.gov.in/pds-caeunfsa.htm}
\BIBentrySTDinterwordspacing

\bibitem[Menon et~al.(2008)Menon, Deolalikar, and
  Bhaskar]{menon2008comparisons}
P.~Menon, A.~Deolalikar, and A.~Bhaskar, \emph{Comparisons of hunger across
  states: India state hunger index}.\hskip 1em plus 0.5em minus 0.4em\relax
  Intl Food Policy Res Inst, 2008.

\end{thebibliography}

\end{document}